\documentclass{article}

\PassOptionsToPackage{numbers,comma,sort&compress}{natbib}

\usepackage[final]{_includes/neurips_2023}

\usepackage[utf8]{inputenc} 
\usepackage[T1]{fontenc}    
\usepackage{hyperref}       
\usepackage{backref}
\usepackage{url}            
\usepackage{booktabs}       
\usepackage{multirow}
\usepackage{amsfonts}       
\usepackage{amsmath}
\usepackage{nicefrac}       
\usepackage{microtype}      
\usepackage{xcolor}         
\usepackage{graphicx}
\usepackage{caption}
\usepackage{subcaption}
\usepackage{wrapfig}

\usepackage{tikz}
\usepackage{pgfplots}
\DeclareUnicodeCharacter{2212}{−}
\usepgfplotslibrary{groupplots,dateplot}
\usetikzlibrary{patterns,shapes.arrows}
\pgfplotsset{compat=newest}

\usepackage{float}
\usepackage{enumitem}
\setlist[itemize]{itemsep=0pt, parsep=0pt, topsep=0pt}

\newcommand{\hallway}{\texttt{hallway}}
\newcommand{\jirpy}{\texttt{jirpy}}

\newcommand{\vocab}{\mathcal{V}}

\definecolor{mpldarkblue}{HTML}{2282bc}
\definecolor{mplcyan}{HTML}{17c6d5}
\definecolor{mplgold}{HTML}{c6c630}
\definecolor{mplgray}{HTML}{8a8a8a}
\definecolor{mplpink}{HTML}{e782c9}
\definecolor{mplbrown}{HTML}{976155}
\definecolor{mplmagenta}{HTML}{9e72c5}
\definecolor{mplred}{HTML}{db2c2d}
\definecolor{mplgreen}{HTML}{32aa32}
\definecolor{mplorange}{HTML}{ff8a0b}

\definecolor{light-gray}{gray}{0.2}
\definecolor{dark-gray}{gray}{0.15}

\hypersetup{
    pdfstartview=FitH,
    pdfpagemode={UseOutlines},
    bookmarksnumbered=true, bookmarksopen=true, colorlinks,
    linkcolor={mplorange},
    citecolor={mpldarkblue},
    urlcolor={mpldarkblue},
    filecolor={mpldarkblue}
}

\usepackage{etoolbox}

\renewcommand*{\backrefalt}[4]{%
    \ifcase #1 %
    No citations.%
    \or
    {\hypersetup{linkcolor=mplgray}
    \textcolor{mplgray}{\hfill \(\rightarrow\) p \hyperpage{#2}}}
    \else
    \textcolor{mplgray}{\hfill \(\rightarrow\) p #2}%
    \fi
}

\newcommand{\dtoprule}{\specialrule{1pt}{0pt}{0.65pt}%
            \specialrule{0.3pt}{0pt}{\belowrulesep}%
            }
\newcommand{\dbottomrule}{\specialrule{0.3pt}{0pt}{0.65pt}%
\specialrule{1pt}{0pt}{\belowrulesep}%
}

\title{Structured World Representations \\ in Maze-Solving Transformers}

\newcommand{\affilCSM}{$^1$} 
\newcommand{\affilICL}{$^2$} 
\newcommand{\affilNIIT}{$^3$} 

\newcommand{\affilspace}{\hspace{0.3cm}}

\newcommand{\authCor}{$^*$ \hspace{-0.5em} }
\newcommand{\authCorText}{\footnotetext[1]{
    Corresponding Author: \href{mailto:mivanits@mines.edu}{mivanits@mines.edu}
}}
\newcommand{\authPri}{$^\dagger$ \hspace{-0.5em} }
\newcommand{\authPriText}{\footnotetext[2]{
    Primary Contributor
}}

\newcommand{\authorbox}[3]{
    \resizebox{#1\columnwidth}{!}{
        \bfseries
        \begin{tabular}{#2}
            #3
        \end{tabular}
    }
}

\author{
    \authorbox{0.65}{ccc}{
        Michael I. Ivanitskiy\authCor \authPri \affilCSM
        & Alex F. Spies\authPri \affilICL \affilNIIT
        & Tilman Räuker\authPri
    } \vspace{0.5em} \\
    \authorbox{1.0}{cccccc}{
        Guillaume Corlouer
        & Chris Mathwin
        & Lucia Quirke
        & Can Rager
        & Rusheb Shah
        & Dan Valentine
    } \vspace{0.5em} \\
    \authorbox{0.6}{ccc}{
        Cecilia Diniz Behn\affilCSM 
        & Katsumi Inoue\affilNIIT 
        & Samy Wu Fung\affilCSM
    }
    \vspace{-1.5em}
}

\newcommand{\makeaffils}{
    \begin{center}
        \parbox{.8\textwidth}{%
          \centering
          \begin{small}
            \affilCSM Colorado School of Mines, Department of Applied Mathematics and Statistics 
            \affilspace
            \affilICL Imperial College London 
            \affilspace
            \affilNIIT National Institute of Informatics, Tokyo
          \end{small}
        }\\
    \end{center}
    \renewcommand*{\thefootnote}{\fnsymbol{footnote}}
    \footnotetext[0]{Code: \href{https://github.com/understanding-search/structured-representations-maze-transformers}{\texttt{github.com/understanding-search/structured-representations-maze-transformers}}}
    \authCorText
    \authPriText
    \renewcommand*{\thefootnote}{\arabic{footnote}}
}

\begin{document}

\maketitle

\makeaffils

\begin{abstract}

Transformer models underpin many recent advances in practical machine learning applications, yet understanding their internal behavior continues to elude researchers. 
Given the size and complexity of these models, forming a comprehensive picture of their inner workings remains a significant challenge. To this end, we set out to understand small transformer models in a more tractable setting: that of solving mazes.
In this work, we focus on the abstractions formed by these models and find evidence for the consistent emergence of structured internal representations of maze topology and valid paths. 
We demonstrate this by showing that the residual stream of only a single token can be linearly decoded to faithfully reconstruct the entire maze. 
We also find that the learned embeddings of individual tokens have spatial structure.
Furthermore, we take steps towards deciphering the circuity of path-following by identifying attention heads (dubbed \textit{adjacency heads}), which are implicated in finding valid subsequent tokens.

\end{abstract}

\section{Introduction}

In recent years, large transformer models have been applied to great effect in various domains, including language modeling, computer vision, and reinforcement learning. The proliferation of such architectures in applied settings has led to increased concern over the generality and robustness of the behaviors they learn. To this end, researchers have begun to study small transformer models on toy tasks to develop a mechanistic understanding of how transformers learn to solve varying classes of problems. The generalizability of findings from toy models to larger scales remains uncertain, but early findings in this direction have given cause for optimism \cite{lieberum2023does}.

The most well-known example of a mechanistic component (a \textit{circuit}) found across many transformers models are induction heads \cite{olsson2022incontext}, which facilitate in-context sequence completion ($A,B,[...],A\to B$) and arise in transformers with at least 2 layers. While induction heads are fairly simple, they form crucial building blocks of more complex circuits
\cite{wang2022interpretability, conmy2023automated}. 
Identifying complete circuits in more complex models is highly labor intensive, but other methods, such as linear probing and the TunedLens \cite{belroseElicitingLatentPredictions2023}, allow researchers to interpret the representations learned by larger models. 
Indeed, recent work \cite{li_othello_2022} found that a GPT-2 model trained on the game of Othello learned to (linearly \cite{nanda_othello_2023}) represent the board state in a way which could be easily intervened upon to change the model's future actions.

\newpage

With the ultimate goal of better understanding how transformer models perform multi-step reasoning in search-like tasks, we apply the interpretability methods to toy models trained to solve maze tasks. In particular, we experiment with autoregressive transformers trained to solve mazes represented as a list of tokens \cite{ivanitskiyConfigurableLibraryGenerating2023}, which constitutes an offline reinforcement learning task with global observations.
By varying the precise configurations of these maze solving tasks, we are able to investigate the conditions under which models tend to learn representations with varying degrees of interpretability and generalizability. 
Additionally, while prior work has found that transformers struggle to perform complex planning tasks \cite{momennejadEvaluatingCognitiveMaps2023}, we find that relatively small ($< 10^7$ parameters) transformers are capable of solving mazes.

We use various interpretability techniques to study our models, finding that the geometry of their embedding space correlates with the spatial structure of the mazes (\autoref{sec:experiments:embed-structure}). 
We find that our highest-performing models form a linear representation of maze connectivity structure, which can be decoded at early layers (\autoref{sec:experiments:worldmodels}). 
Lastly, we identify specific attention heads that condition over valid neighbors for a given state, implicating them in path-following behavior (\autoref{sec:experiments:dla} and \autoref{sec:experiments:tunedlens}). 

By performing these analyses across models and at different stages in training, we find evidence for grokking-like transitions during training, in which a model's ability to generalize improves rapidly \cite{nandaProgressMeasuresGrokking2023}. These increases in generalization performance coincide with the times at which models' internal representations of the maze become more linearly decodable, suggesting that a structured internal representation improves their ability to systematically solve mazes (\autoref{sec:experiments:grokking}).

\section{Experimental Setting}

\subsection{Datasets}
\label{sec:dataset}

\begin{figure}
     \centering
     \begin{subfigure}[b]{0.61\textwidth}
         \centering
         \begin{minipage}{\textwidth}
            \fontsize{6pt}{6pt}\selectfont
            \setlength{\fboxsep}{2pt}
            \input{figures/dataset/outputs-tokens-colored-shortened.tex} 
         \end{minipage}
         \vspace{0pt}
         \setlength{\fboxsep}{1pt}
         \vspace{0pt}
         \caption{An example training sequence with four parts. \textbf{1:} The \colorbox[RGB]{ 217,210,233 }{adjacency list} describes the connectivity of the maze (the order of connections is randomized, ellipses represent omitted connection pairs). \textbf{2,3:} The \colorbox[RGB]{ 217,234,211 }{origin} and \colorbox[RGB]{ 234,209,220 }{target} specify where the path should begin and end, respectively. \textbf{4:} The \colorbox[RGB]{ 207,226,243 }{path} itself is the shortest sequence of coordinates from the origin to the target. For a ``rollout,'' we provide everything up to (and not including) the \texttt{<PATH\_START>} token and generate a sequence via \texttt{argmax} sampling until the \texttt{<PATH\_END>} token is produced. For single-token tasks (see \autoref{fig:task-desc}), we provide a partially complete path and consider only the logits over the immediate next token.}
         \label{fig:methodology.maze_task.visualization}
     \end{subfigure}
     \hfill
     \begin{subfigure}[b]{0.35\textwidth}
         \centering
         \includegraphics[width=0.8\textwidth]{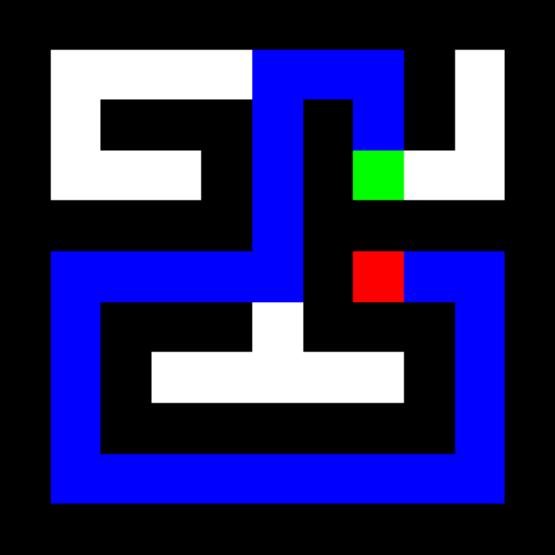}
         \vspace{0pt}
         \caption{Visual representation of the maze defined from the tokens on the left. The origin is indicated in green, the target in red, and the correct path in blue.}
         \label{fig:methodology.maze_task.sequence}
     \end{subfigure}
     \caption{Tokenization scheme and visualization of our shortest-path maze tasks.}
    \label{fig:methodology.maze_task}
\end{figure}

We use the \texttt{maze-dataset} library \cite{ivanitskiyConfigurableLibraryGenerating2023} to generate a variety of mazes and convert them into formats suitable for a text-based autoregressive transformer. Starting with an $n \times n$ lattice, we generate paths using a variety of algorithms. The resulting mazes are converted into tokenized representations, shown in \autoref{fig:methodology.maze_task.sequence}, which are used to train our models. The vocabulary consists of coordinate tokens and various special tokens used to connect coordinates and delimit different parts of the maze and solution description. While \texttt{maze-dataset} provides a variety of maze generation algorithms, filters, and configuration parameters, in our interpretability experiments, we focus on mazes generated via 1) Randomized Depth First Search (RDFS), which generates acyclic spanning trees; 2) ``forkless'' mazes consisting of a sparse tree where each node has at most two connections; 3) Randomized Depth First Search with percolation (pRDFS), which starts with RDFS, but then a \texttt{OR} is performed with a maze where adjacent connections have probability $p=0.1$ of occurring, thus creating mazes which may have cycles.

\newpage

\subsection{Models and training}
\label{sec:models-training}

All models analyzed are autoregressive decoder-only models, identical to the GPT architecture. While extensive sweeps were performed over hyperparameters, we focused our experiments on two trained models. The first, denoted \hallway, was trained only on ``forkless'' mazes and is a smaller model with approximately $1.2$M parameters. The second model, \jirpy, has approximately $9.6$M parameters and was trained on sparsely connected mazes of varying sizes with multiple forking points (see \autoref{sec:dataset}).\footnote{Chosen as it was the most performant of the models trained to solve complex mazes. See \autoref{fig:app:sweep_performance}.} These two models trade off interpretability and task complexity, with the \hallway \ task allowing for a simpler model, while the task for \jirpy \ requires decision making at each forking point, potentially yielding a more complex maze representation. Full hyperparameters for our models can be found in the appendix.

Models were trained to perform next-token prediction on a dataset of randomly generated mazes and paths.\footnote{\jirpy\ received gradients only from tokens in the path (including special delimiters), while \hallway\ received gradients from the entire sequence.} At inference time, the models are prompted with a complete adjacency list and path specification (i.e., all tokens up to \texttt{<PATH\_START>}) and rolled out until they yield a \texttt{<PATH\_END>} token. It is worth noting that we do not impose any constraints on the validity of a model's output, so a poorly trained model may output nonsensical paths consisting of special tokens or disconnected coordinates.
 
\section{Experiments}
\label{sec:experiments}

To understand our trained maze-solving transformers' behavior and internal representations, we favor a post-hoc interpretability approach \cite{rauker2023transparent}.
We begin with behavioral experiments on maze-solving trajectories and assess initial path predictions. 
Next, we explore the embedding space for spatial token relationships and use direct logit attribution \cite{wang2022interpretability, lieberum2023does} to pinpoint model components sensitive to specific sub-tasks. 
Through linear probes on the residual stream, we decode the presence or absence of walls, revealing structured representations. 
Lastly, we analyze training metrics to investigate the relationship between the emergence of structured representations and improved generalization performance. Collectively, these experiments shed light on how our transformers adeptly solve mazes.

\subsection{Behavioral Experiments}
\label{sec:experiments:behavioral-experiments}

\begin{figure}
    \centering
    \includegraphics[width=\textwidth]{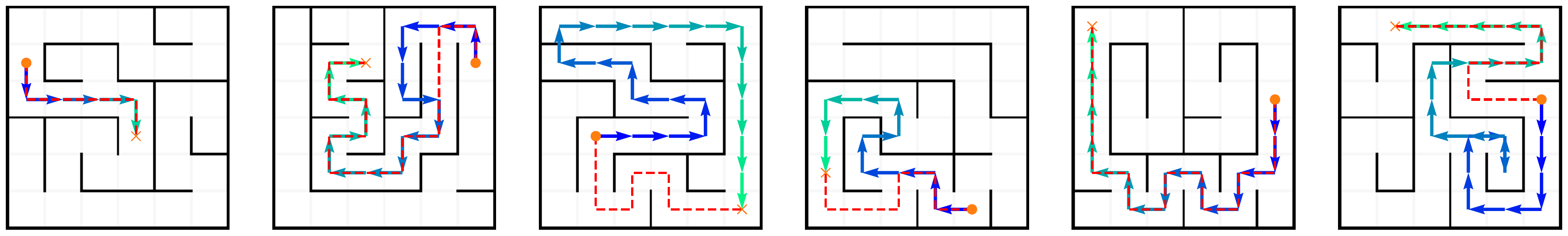}
    \includegraphics[width=\textwidth]{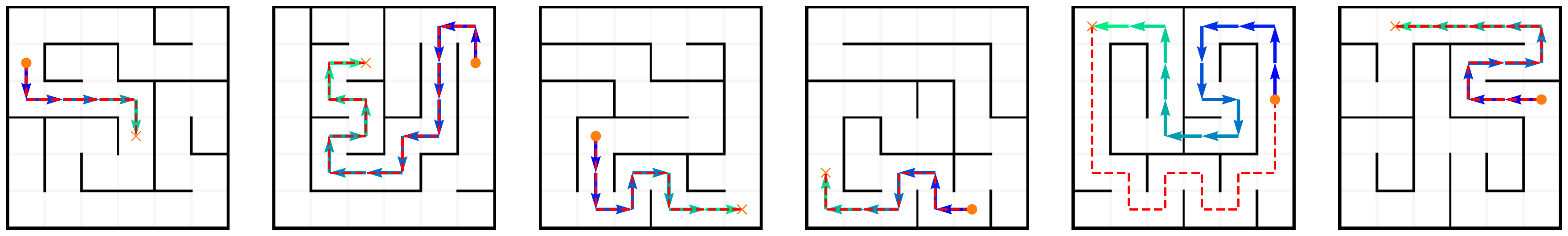}
    \caption{Example generations of \hallway \ model (top row) and \jirpy \ model (bottom row) on a random sample of held-out RDFS mazes (outside the training distribution of the \hallway \ model). The correct path is marked as a red dashed line, with \textcolor{red}{$\bullet$} at the starting position and \textcolor{red}{$\texttt{x}$} at the target position. For clarity, generated paths fade from blue to green. Note that both models often violate constraints, such as by passing through walls, and reach the target despite being at a dead end. Further example generations can be found with our codebase.}
    \label{fig:rollouts}
\end{figure}

Although several evaluation metrics are computed during the training process, we found visual inspection of generated paths to be useful. Several example rollouts are provided in \autoref{fig:rollouts}.

To facilitate the isolation of specific sub-components of our transformers, which are implicated in certain behaviors, we identify sub-``tasks'' of maze solving, which consist of predicting a single token. We describe several such tasks in \autoref{fig:task-desc}. For each of these, the prompt given to the model consists of all context tokens up to and not including the targeted token. Of particular note in our experiments is the qualitative observation that the models tend to reach the goal at the conclusion of their generations but often violate the constraints in the process\footnote{I.e. its output sequence is often of the form ``\texttt{ [...invalid path...], (goal), <PATH\_END>}''.}, as shown in \autoref{fig:rollouts} and \autoref{tab:task-accuracies}.

\begin{figure}
    \centering
    \begin{tabular}{cccccc}
        \colorbox[RGB]{162,209,191}{\texttt{<PATH\_START>}} &
        \colorbox[RGB]{217,234,211}{\texttt{(1,3)}} & 
        \colorbox[RGB]{217,210,233}{\texttt{(0,3)}} & 
        \colorbox[RGB]{207,226,243}{\texttt{(0,2) [...] (2,4)}} & 
        \colorbox[RGB]{230,200,220}{\texttt{(2,3)}} & 
        \colorbox[RGB]{201,179,131}{\texttt{<PATH\_END>}} \\
    \end{tabular}
    \caption{
        \footnotesize
        \setlength{\fboxsep}{1.5pt}
        Tasks used to assess model performance and their relative locations within a path prediction. From left to right, the target tokens are:
        \colorbox[RGB]{162,209,191}{\texttt{path\_start}}, 
        \colorbox[RGB]{217,234,211}{\texttt{origin\_after\_path\_start}}, 
        \colorbox[RGB]{217,210,233}{\texttt{first\_path\_choice}}, 
        \colorbox[RGB]{207,226,243}{\texttt{rand\_path\_token\_nonend}},  
        \colorbox[RGB]{230,200,220}{\texttt{final\_before\_path\_end}}, 
        \colorbox[RGB]{201,179,131}{\texttt{path\_end}}.
        Notably, for hallway-type tasks, the \texttt{first\_path\_choice} task is the only task that requires anything other than simple following of a path and recognition of the origin and target. A \texttt{rand\_path\_token} task is also included, which is similar to \texttt{rand\_path\_token\_nonend} in that one of several tokens is selected at random, but for the latter, this pool of possible tokens does not exclude endpoints. Performance on these tasks is shown in \autoref{tab:task-accuracies}.
    }
    \label{fig:task-desc}
\end{figure}

In \autoref{tab:task-accuracies}, we note that on out-of-distribution pRDFS mazes, both models generalize fairly well. We observe that performance on the \texttt{first\_path\_choice} task is consistently the lowest. Performance of \hallway\ on $6 \times 6$ mazes is slightly lower than on larger mazes (see \autoref{tab:task-accuracies-g7-APPENDIX}), possibly due to the short prompt length being out-of-distribution.

\begin{table}
    \centering
    { \footnotesize 
        \begin{tabular}{l|rrrrrr}
\toprule
             \multicolumn{1}{r}{dataset:}   & \multicolumn{2}{c}{forkless} & \multicolumn{2}{c}{RDFS} & \multicolumn{2}{c}{pRDFS} \\
             \multicolumn{1}{r}{model:}     & \hallway &   \jirpy & \hallway &    \jirpy &  \hallway &   \jirpy \\
\midrule
\midrule
                  exactly correct rollouts &  \textbf{38.3\%} &   \textbf{38.7\%} &   24.2\% &    82.4\% &    24.2\% &    70.7\% \\
                            valid rollouts &  \textbf{54.3\%} &   \textbf{53.5\%} &   37.5\% &    84.0\% &    49.6\% &    87.1\% \\
              rollouts with target reached &  \textbf{87.1\%} &   \textbf{64.5\%} &   94.5\% &    99.2\% &    92.6\% &   100.0\% \\
\midrule
                     \texttt{path\_start}  & 100.0\% &  100.0\% &  100.0\% &   100.0\% &   100.0\% &   100.0\% \\
       \texttt{origin\_after\_path\_start} &  91.0\% &   86.7\% &  100.0\% &   100.0\% &   100.0\% &   100.0\% \\
              \texttt{first\_path\_choice} &  \textbf{71.5\%} &   \textbf{66.4\%} &   \textbf{67.2\%} &    \textbf{86.7\%} &    \textbf{66.4\%} &    \textbf{84.4\%} \\
                \texttt{rand\_path\_token} &  93.0\% &   87.1\% &   90.2\% &    98.0\% &    84.0\% &    94.5\% \\
 \texttt{rand\_path\_token\_nonend} &  97.3\% &   89.8\% &   92.2\% &    99.2\% &    84.4\% &    97.3\% \\
         \texttt{final\_before\_path\_end} &  95.7\% &   85.9\% &   93.4\% &   100.0\% &    84.4\% &   100.0\% \\
                        \texttt{path\_end} &  86.7\% &   71.5\% &  100.0\% &    99.6\% &   100.0\% &   100.0\% \\
\bottomrule
\end{tabular}

    }
    \vspace{0.8em}
    \caption{\textbf{Setup:} Performance across tasks of the \hallway \ model and \jirpy \ model assessed on held-out $6 \times 6$ forkless, RDFS, and pRDFS mazes (See \autoref{sec:dataset} and \cite{ivanitskiyConfigurableLibraryGenerating2023}). All values are binary, since we perform a single rollout per maze, and score for a task is \texttt{argmax(logits) == correct\_token}. \textbf{Tasks:}  The first group of metrics deals with sequence generations or ``rollouts,'' as detailed in \autoref{sec:dataset}. A rollout is ``exactly correct'' if no deviations from the shortest path occur (pRDFS mazes may not have a unique shortest path, and thus the provided values are a \emph{lower bound}). A rollout is ``valid'' if it obeys the topology of the maze (no wall jumps), but backtracking is permitted. A rollout is considered to have reached the target if the final coordinate token is the target token. The second group of single-token tasks used to assess performance are detailed in \autoref{fig:task-desc}. Data for $7 \times 7$ mazes is provided in \autoref{sec:performance-APPENDIX}.}
    \label{tab:task-accuracies}
\end{table}

\subsection{Emergent Structure in the Embedding Space}\label{sec:experiments:embed-structure}

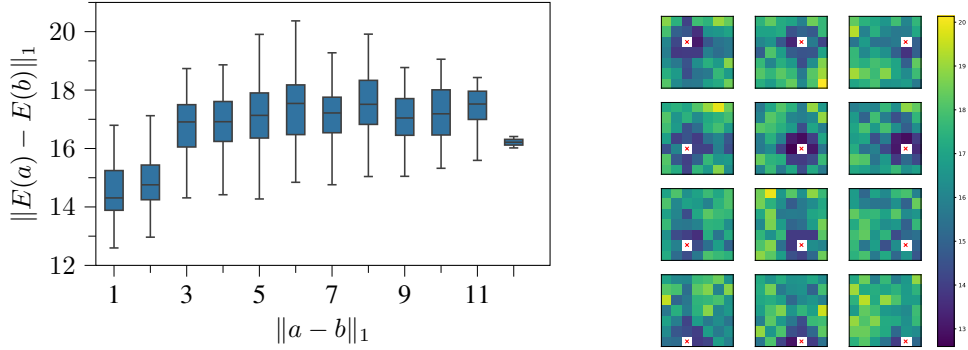
\begin{figure}
    \centering
    \begin{minipage}[t]{0.50\textwidth}
\begin{tikzpicture}

\definecolor{darkgray176}{RGB}{176,176,176}
\definecolor{darkslategray63}{RGB}{63,63,63}
\definecolor{steelblue49115161}{RGB}{49,115,161}

\begin{axis}[
height=2in, width=3in,
tick align=outside,
tick pos=left,
x grid style={darkgray176},
xlabel={$\Vert a - b \Vert_1$},
xmin=-0.5, xmax=12,
xtick style={color=black},
xtick={0,2,4,6,8,10},
xticklabels={1,3,5,7,9,11},
minor x tick num=1,
extra x ticks={11},
extra x tick style={xticklabel=\empty},
y grid style={darkgray176},
ylabel={$\Vert E(a) - E(b) \Vert_1$},
ymin=12.0, ymax=21.0,
minor y tick num=1,
ytick style={color=black},
outer sep=0pt,
xlabel style={yshift=1em},
ylabel style={yshift=-0.2em}
]
\path [draw=darkslategray63, fill=steelblue49115161, semithick]
(axis cs:-0.25,13.8853633739782)
--(axis cs:0.25,13.8853633739782)
--(axis cs:0.25,15.2452638835821)
--(axis cs:-0.25,15.2452638835821)
--(axis cs:-0.25,13.8853633739782)
--cycle;
\path [draw=darkslategray63, fill=steelblue49115161, semithick]
(axis cs:0.75,14.2466728058062)
--(axis cs:1.25,14.2466728058062)
--(axis cs:1.25,15.434527534846)
--(axis cs:0.75,15.434527534846)
--(axis cs:0.75,14.2466728058062)
--cycle;
\path [draw=darkslategray63, fill=steelblue49115161, semithick]
(axis cs:1.75,16.0521065542853)
--(axis cs:2.25,16.0521065542853)
--(axis cs:2.25,17.5004779411829)
--(axis cs:1.75,17.5004779411829)
--(axis cs:1.75,16.0521065542853)
--cycle;
\path [draw=darkslategray63, fill=steelblue49115161, semithick]
(axis cs:2.75,16.2421603152761)
--(axis cs:3.25,16.2421603152761)
--(axis cs:3.25,17.608891440148)
--(axis cs:2.75,17.608891440148)
--(axis cs:2.75,16.2421603152761)
--cycle;
\path [draw=darkslategray63, fill=steelblue49115161, semithick]
(axis cs:3.75,16.3580248670623)
--(axis cs:4.25,16.3580248670623)
--(axis cs:4.25,17.9023113381409)
--(axis cs:3.75,17.9023113381409)
--(axis cs:3.75,16.3580248670623)
--cycle;
\path [draw=darkslategray63, fill=steelblue49115161, semithick]
(axis cs:4.75,16.4760989408824)
--(axis cs:5.25,16.4760989408824)
--(axis cs:5.25,18.1775068740244)
--(axis cs:4.75,18.1775068740244)
--(axis cs:4.75,16.4760989408824)
--cycle;
\path [draw=darkslategray63, fill=steelblue49115161, semithick]
(axis cs:5.75,16.5395179956977)
--(axis cs:6.25,16.5395179956977)
--(axis cs:6.25,17.7564820022671)
--(axis cs:5.75,17.7564820022671)
--(axis cs:5.75,16.5395179956977)
--cycle;
\path [draw=darkslategray63, fill=steelblue49115161, semithick]
(axis cs:6.75,16.825512707117)
--(axis cs:7.25,16.825512707117)
--(axis cs:7.25,18.3329722944763)
--(axis cs:6.75,18.3329722944763)
--(axis cs:6.75,16.825512707117)
--cycle;
\path [draw=darkslategray63, fill=steelblue49115161, semithick]
(axis cs:7.75,16.4526583119587)
--(axis cs:8.25,16.4526583119587)
--(axis cs:8.25,17.7082145048771)
--(axis cs:7.75,17.7082145048771)
--(axis cs:7.75,16.4526583119587)
--cycle;
\path [draw=darkslategray63, fill=steelblue49115161, semithick]
(axis cs:8.75,16.4626767859445)
--(axis cs:9.25,16.4626767859445)
--(axis cs:9.25,18.0086692394689)
--(axis cs:8.75,18.0086692394689)
--(axis cs:8.75,16.4626767859445)
--cycle;
\path [draw=darkslategray63, fill=steelblue49115161, semithick]
(axis cs:9.75,16.9947069542832)
--(axis cs:10.25,16.9947069542832)
--(axis cs:10.25,17.9612823321368)
--(axis cs:9.75,17.9612823321368)
--(axis cs:9.75,16.9947069542832)
--cycle;
\path [draw=darkslategray63, fill=steelblue49115161, semithick]
(axis cs:10.75,16.1176965087943)
--(axis cs:11.25,16.1176965087943)
--(axis cs:11.25,16.3129950432049)
--(axis cs:10.75,16.3129950432049)
--(axis cs:10.75,16.1176965087943)
--cycle;
\addplot [semithick, darkslategray63]
table {%
0 13.8853633739782
0 12.5961953881197
};
\addplot [semithick, darkslategray63]
table {%
0 15.2452638835821
0 16.7951860975008
};
\addplot [semithick, darkslategray63]
table {%
-0.125 12.5961953881197
0.125 12.5961953881197
};
\addplot [semithick, darkslategray63]
table {%
-0.125 16.7951860975008
0.125 16.7951860975008
};
\addplot [semithick, darkslategray63]
table {%
1 14.2466728058062
1 12.9655751143582
};
\addplot [semithick, darkslategray63]
table {%
1 15.434527534846
1 17.1232053181157
};
\addplot [semithick, darkslategray63]
table {%
0.875 12.9655751143582
1.125 12.9655751143582
};
\addplot [semithick, darkslategray63]
table {%
0.875 17.1232053181157
1.125 17.1232053181157
};
\addplot [semithick, darkslategray63]
table {%
2 16.0521065542853
2 14.3102743688505
};
\addplot [semithick, darkslategray63]
table {%
2 17.5004779411829
2 18.7366737478878
};
\addplot [semithick, darkslategray63]
table {%
1.875 14.3102743688505
2.125 14.3102743688505
};
\addplot [semithick, darkslategray63]
table {%
1.875 18.7366737478878
2.125 18.7366737478878
};
\addplot [semithick, darkslategray63]
table {%
3 16.2421603152761
3 14.4175735963508
};
\addplot [semithick, darkslategray63]
table {%
3 17.608891440148
3 18.8639378646039
};
\addplot [semithick, darkslategray63]
table {%
2.875 14.4175735963508
3.125 14.4175735963508
};
\addplot [semithick, darkslategray63]
table {%
2.875 18.8639378646039
3.125 18.8639378646039
};
\addplot [semithick, darkslategray63]
table {%
4 16.3580248670623
4 14.2728807174135
};
\addplot [semithick, darkslategray63]
table {%
4 17.9023113381409
4 19.9067178193945
};
\addplot [semithick, darkslategray63]
table {%
3.875 14.2728807174135
4.125 14.2728807174135
};
\addplot [semithick, darkslategray63]
table {%
3.875 19.9067178193945
4.125 19.9067178193945
};
\addplot [semithick, darkslategray63]
table {%
5 16.4760989408824
5 14.8434002918657
};
\addplot [semithick, darkslategray63]
table {%
5 18.1775068740244
5 20.3687717046123
};
\addplot [semithick, darkslategray63]
table {%
4.875 14.8434002918657
5.125 14.8434002918657
};
\addplot [semithick, darkslategray63]
table {%
4.875 20.3687717046123
5.125 20.3687717046123
};
\addplot [semithick, darkslategray63]
table {%
6 16.5395179956977
6 14.7599147340516
};
\addplot [semithick, darkslategray63]
table {%
6 17.7564820022671
6 19.2761099518393
};
\addplot [semithick, darkslategray63]
table {%
5.875 14.7599147340516
6.125 14.7599147340516
};
\addplot [semithick, darkslategray63]
table {%
5.875 19.2761099518393
6.125 19.2761099518393
};
\addplot [semithick, darkslategray63]
table {%
7 16.825512707117
7 15.0408815378323
};
\addplot [semithick, darkslategray63]
table {%
7 18.3329722944763
7 19.9147282083286
};
\addplot [semithick, darkslategray63]
table {%
6.875 15.0408815378323
7.125 15.0408815378323
};
\addplot [semithick, darkslategray63]
table {%
6.875 19.9147282083286
7.125 19.9147282083286
};
\addplot [semithick, darkslategray63]
table {%
8 16.4526583119587
8 15.0482227395987
};
\addplot [semithick, darkslategray63]
table {%
8 17.7082145048771
8 18.7712432030821
};
\addplot [semithick, darkslategray63]
table {%
7.875 15.0482227395987
8.125 15.0482227395987
};
\addplot [semithick, darkslategray63]
table {%
7.875 18.7712432030821
8.125 18.7712432030821
};
\addplot [semithick, darkslategray63]
table {%
9 16.4626767859445
9 15.3229888981441
};
\addplot [semithick, darkslategray63]
table {%
9 18.0086692394689
9 19.0561846718192
};
\addplot [semithick, darkslategray63]
table {%
8.875 15.3229888981441
9.125 15.3229888981441
};
\addplot [semithick, darkslategray63]
table {%
8.875 19.0561846718192
9.125 19.0561846718192
};
\addplot [semithick, darkslategray63]
table {%
10 16.9947069542832
10 15.5934676951729
};
\addplot [semithick, darkslategray63]
table {%
10 17.9612823321368
10 18.4294692674885
};
\addplot [semithick, darkslategray63]
table {%
9.875 15.5934676951729
10.125 15.5934676951729
};
\addplot [semithick, darkslategray63]
table {%
9.875 18.4294692674885
10.125 18.4294692674885
};
\addplot [semithick, darkslategray63]
table {%
11 16.1176965087943
11 16.0200472415891
};
\addplot [semithick, darkslategray63]
table {%
11 16.3129950432049
11 16.4106443104101
};
\addplot [semithick, darkslategray63]
table {%
10.875 16.0200472415891
11.125 16.0200472415891
};
\addplot [semithick, darkslategray63]
table {%
10.875 16.4106443104101
11.125 16.4106443104101
};
\addplot [semithick, darkslategray63]
table {%
-0.25 14.3089029366383
0.25 14.3089029366383
};
\addplot [semithick, darkslategray63]
table {%
0.75 14.7573162942135
1.25 14.7573162942135
};
\addplot [semithick, darkslategray63]
table {%
1.75 16.9098038851807
2.25 16.9098038851807
};
\addplot [semithick, darkslategray63]
table {%
2.75 16.9172350766312
3.25 16.9172350766312
};
\addplot [semithick, darkslategray63]
table {%
3.75 17.1340310644009
4.25 17.1340310644009
};
\addplot [semithick, darkslategray63]
table {%
4.75 17.5424086888088
5.25 17.5424086888088
};
\addplot [semithick, darkslategray63]
table {%
5.75 17.2169249798171
6.25 17.2169249798171
};
\addplot [semithick, darkslategray63]
table {%
6.75 17.5127020963118
7.25 17.5127020963118
};
\addplot [semithick, darkslategray63]
table {%
7.75 17.044514516776
8.25 17.044514516776
};
\addplot [semithick, darkslategray63]
table {%
8.75 17.1879512511077
9.25 17.1879512511077
};
\addplot [semithick, darkslategray63]
table {%
9.75 17.5225474389736
10.25 17.5225474389736
};
\addplot [semithick, darkslategray63]
table {%
10.75 16.2153457759996
11.25 16.2153457759996
};
\end{axis}


\end{tikzpicture}
    \end{minipage}
    \hspace{0.6in}
    \includegraphics[width=0.3\textwidth]{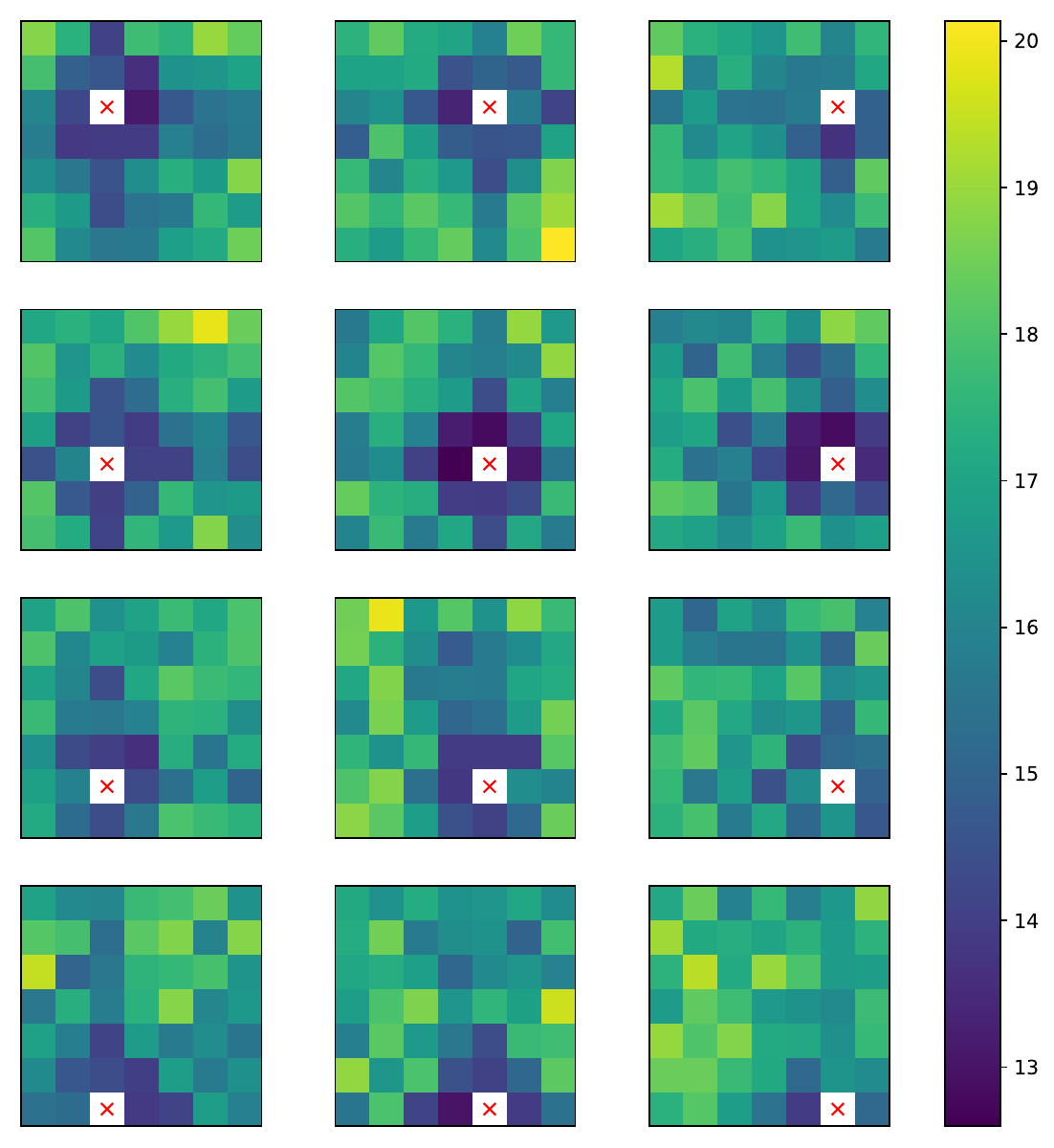}
    
    \caption{Structure of coordinate token embeddings for the \hallway \ model. Given two coordinates $a,b$ and the embeddings of their corresponding tokens $E(a),E(b)$ we observe the relationship between the Manhattan ($1$-norm) distances. Note that all coordinates have orthogonal vocabulary vectors, and the embeddings are learned. \textbf{Left:} Correlation between coordinate distance and embedding distance. \textbf{Right:} Given the embedding of the coordinate at the \textcolor{red}{\texttt{x}}, Manhattan distance to the other tokens on the grid is displayed. \textbf{Note:} full data for all models can be seen in \autoref{sec:appendix:embed}.}
    \label{fig:embed-structure}
\end{figure}

As in other language models, each token in the vocabulary corresponds to a unique orthogonal unit vector. In our experiments, each coordinate on the lattice has a single corresponding token. The embedding layer of our models maps each vocabulary vector from an input sequence to a dense vector in $\mathbb{R}^{d_{\texttt{model}}}$. Since each vocabulary vector is orthogonal, no spatial structure is encoded into the model directly; however, a spatial structure emerges after we train the model. In particular, we note that a correlation between the coordinate distance and distance between embedding vectors emerges for short distances (\autoref{fig:embed-structure}). Note that proximity of tokens in the sequences alone is not enough to allow this behavior to be learned due to the randomization of the adjacency list. 

\subsection{Direct Logit Attribution}
\label{sec:experiments:dla}
\begin{figure}
    \centering
    \includegraphics[width=0.24\textwidth]{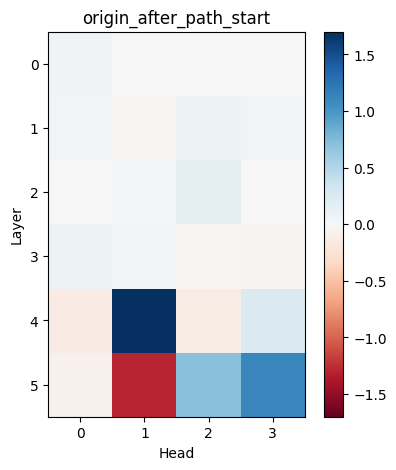}
    \includegraphics[width=0.24\textwidth]{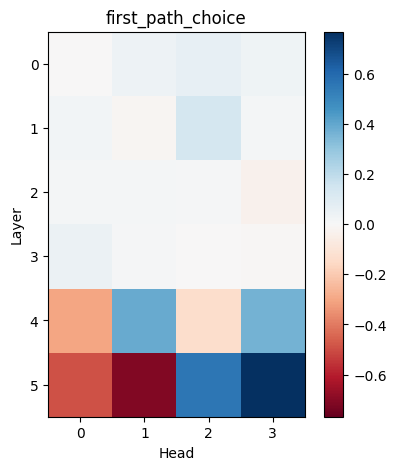}
    \includegraphics[width=0.24\textwidth]{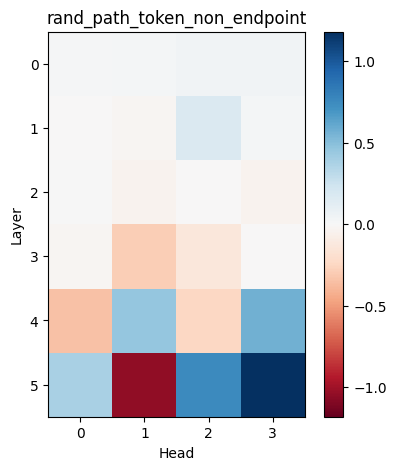}
    \includegraphics[width=0.24\textwidth]{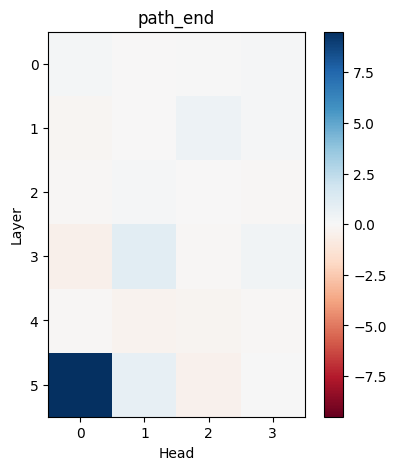}
    \caption{DLA of the \hallway \ model across a subset of tasks, on held-out samples from the training distribution. The numerical value is the contribution of a given attention head to the ``correct'' direction in the residual stream. Note that only for \texttt{first\_path\_choice} must the model do anything besides path-following, and this is shown in the performance statistics of \autoref{tab:task-accuracies}.}
    \label{fig:dla-rand_nonend}
\end{figure}

To investigate path-following behavior, we utilize direct logit attribution (DLA) \cite{wang2022interpretability, lieberum2023does}, which measures the direct contribution of an isolated component of the network (e.g., an attention head) to a given set of forward passes. 
We utilize the tasks defined in \autoref{sec:experiments:behavioral-experiments} for this correlational analysis (\autoref{fig:dla-rand_nonend}). 
Specifically, we compute the contribution $C_{l,h}$ of head $h$ at layer $l$ to the probability assigned by the model to the correct next token. 
We do this by empirically estimating (over samples $(p, c) \in \mathcal{D}$) the dot product of the output\footnote{Note that the application of LayerNorm is done to match the actual scaling at layer $l$, see \texttt{ActivationCache.apply\_ln\_to\_stack()} in \cite{nanda2022transformerlens}.} 
of that head $R_{l,h}(p)$ with the difference between the embedding of the correct token $E(c)$ and some reference embedding $r(c)$.
$$
    C_{l,h} = \frac{\scriptstyle 1}{\scriptstyle |\mathcal{D}|}
        \sum_{(p,c) \in \mathcal{D}}\big[
        \texttt{LayerNorm}\left(
            R_{l,h}(p)
        \right) \cdot (E(c) - r(c))
    \big]
    \qquad
    \text{where}
    \qquad
    r(c) = 
        \frac{\scriptstyle 1}{\scriptstyle |\vocab|-1} 
        \sum_{
            t \in \vocab \setminus c
        }
        E(t)
$$

Where $\vocab$ is the set of vocabulary vectors, we compute\footnote{In a manner which is not common practice, to our knowledge.} the reference embedding as a mean of the embeddings of all tokens except $c$.
In this work, the DLA analysis serves to locate attention heads of interest, which we subsequently investigate. Our future work will include ablations and other interventions on model architecture to establish causal relationships between these attention heads and path-following performance.

Upon investigation of attention placed on tokens as a function of their distance from the current token, we find that Layer 5, Head 0 simply places attention on the recent occurrences of the current coordinate token. This is throughout the whole sequence, but primarily between the target specification tokens. As such, its lack of involvement in \texttt{origin\_after\_path\_start} becomes clear since the current token, in that case, would be the \texttt{<PATH\_START>} token and thus not a coordinate token.

\vspace{1em}

Also of interest is Layer 1, Head 2. We find that consistently across tasks, this head places attention on the \texttt{<TARGET\_END>} token. We hypothesize that this head is a component of an induction head \cite{olsson2022incontext} but operating in reverse -- a later head likely attends to the token before \texttt{<TARGET\_END>} to find the target. This information may then be used to inform the model's choices of which path to take at forks, as well as identifying when the path is concluded.

\vspace{2em}

\begin{figure}
    \centering
    \includegraphics[width=\textwidth]{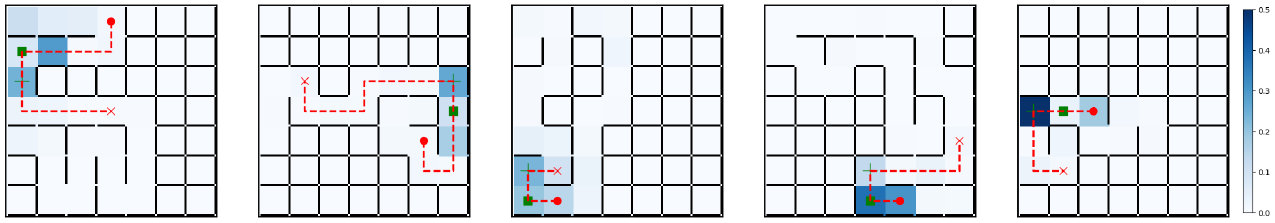}
    
    \caption{Attention of Layer 5, Head 3, on the \texttt{rand\_path\_token\_nonend} task. Attention is displayed over the maze positions for five random held-out mazes. Blue shading is attention weight, true path is red dashed line from \textcolor{red}{$\bullet$} to \textcolor{red}{$\texttt{x}$}, current position is {\color[RGB]{0,140,0}\rule{1ex}{1ex}}.}
    \label{fig:adj-head-maze}
\end{figure}

\vspace{2em}

\begin{figure}
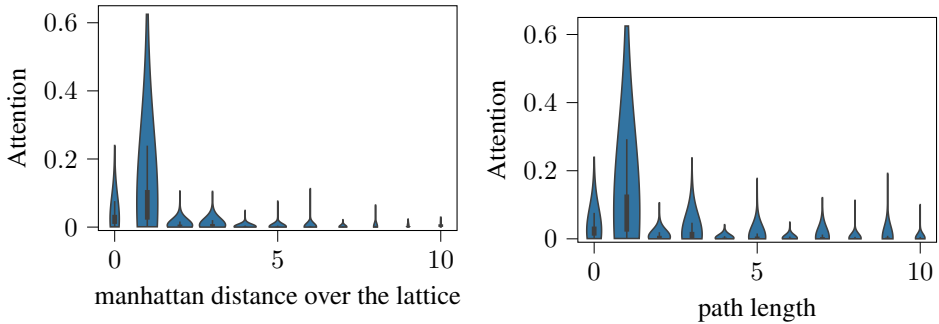

    \centering

    \begin{minipage}{0.45\textwidth}
        \input{figures/dla/attn/dist-corr-lattice-L5-H3}
    \end{minipage}
    \begin{minipage}{0.45\textwidth}
        \input{figures/dla/attn/dist-corr-topology-L5-H3}
    \end{minipage}
    
    \caption{Attention of Layer 5, Head 3, on the \texttt{rand\_path\_token\_nonend} task: violin plots of attention as a function of the distance between the current node and the node being attended to. $x$-axis on the left is the pure manhattan distance between the notes, while $x$-axis on the right is the path length between the nodes. Sample size $n=200$. Note that while on the right, attention is overwhelmingly applied to nodes path length 1 away, some attention is applied to nodes at odd path lengths away because a node adjacent to the lattice will always be an odd path length away.}
    \label{fig:adj-head-corr}
\end{figure}

\vspace{2em}

As observed in \autoref{fig:adj-head-maze} and \autoref{fig:adj-head-corr}, the attention head at layer 5, head 3, which we term an \emph{Adjacency Head}, consistently attends to tokens of path length 1 from the current position and thereby learns to \emph{respect the maze's topology}. This differs from the results of \autoref{sec:experiments:embed-structure} in that the embedding map, since it processes each token individually, can only correlate vectors that are close on the lattice (shared across all training runs) and cannot see the topology which can only be learned in-context. 

\newpage

\subsection{Learned internal representations}
\label{sec:experiments:worldmodels}

To assess whether the models learn to internally represent mazes, we follow the approach in \cite{nanda_othello_2023} and train a set of linear probes to predict the ground-truth maze structure from a single latent vector. In particular, for a maze with $m\times m$ positions, we train $n_{\texttt{layers}} \times m \times m \times 4$ probes $p$ on residual stream activations $R_l(t)$ collected across many rollouts, such that
$$
     \left[
        R_l(t) 
        \cdot p_l(x,y,\texttt{dir})
    \right] > 0.5 
    = \texttt{wall}
        (x,y,\texttt{dir})
$$
where the token, $t$, is fixed and all layers, $l$, are considered. In essence, if the dot product between a particular direction $\texttt{dir}$ probe with $R_l(t) $ exceeds 0.5, then this should reflect the presence of a wall in the input maze at that particular probing location. For all experiments, we take the token, $t$, to be \texttt{<PATH\_START>}, as it is the final token presented in each sequence at test time and will have seen all previous tokens.

\begin{figure}
     \centering
     \makebox[\textwidth][c]{

     \hfill
     \begin{subfigure}[b]{0.3\textwidth}
         \centering
         \includegraphics[height=4cm]{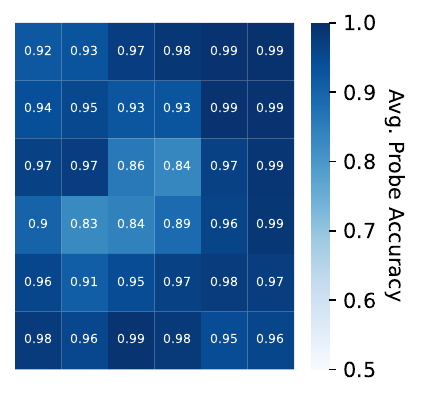}
         \caption{Probe set accuracy across maze positions for layer 2 (for which \jirpy\ yielded the most accurate probes).}
         \label{fig:linear_probing_acc_jerpkipj_PATH_START_single_layer}
     \end{subfigure}
     \hfill
     \begin{subfigure}[b]{0.65\textwidth}
         \centering
         \includegraphics[height=4cm]{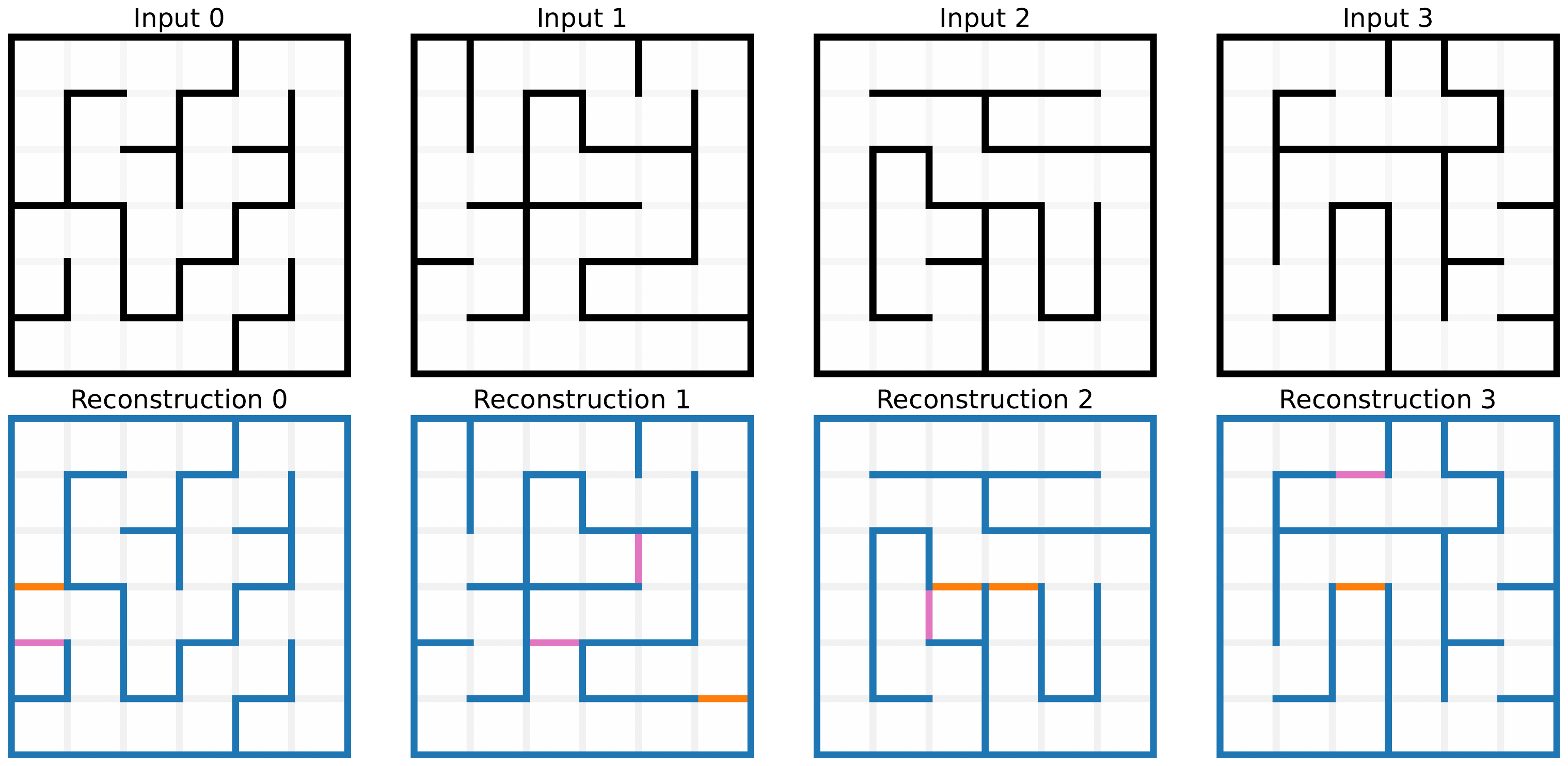}
         \caption{Four random examples of mazes reconstructed using the probes. Reconstructions are made from a set of probes using a single latent embedding of the \texttt{<PATH\_START>} token at layer 2. Wall colors indicate that thresholded probes \textcolor{mpldarkblue}{Correctly Predicted}, \textcolor{mplorange}{Omitted} or \textcolor{mplpink}{Added} a wall.}
         \label{fig:jerpkipj_PATH_START_reconstruct_maze_multi_example_layer_2}
     \end{subfigure}
     }
        \caption{Analysis of Linear Probes applied to the \texttt{<PATH\_START>} token for the \jirpy \ model.}
        \label{fig:experiments.probes.accuracies}
\end{figure}

By looking at the variation in probing accuracy across layers and throughout training, we can understand how the formation of the world model occurs and potentially contributes to the model's performance. We focus our discussion here on \jirpy \ as it was the most performant model, both in terms of solving mazes and yielding the best set of probes (see \autoref{fig:app:linear_probing_all} for the results on all layers). In \autoref{fig:experiments.probes.accuracies} we show the examples of mazes decoded at layer 2 with a set of probes that achieved the highest accuracy. \autoref{fig:experiments.probes.layerwise} shows that the maze representation was already learned by the second layer (more examples are shown in \autoref{fig:app:reco_mazes_more_1} and \autoref{fig:app:reco_mazes_more_2}). Results of sweeps across different transformers are shown in \autoref{fig:app:sweep_performance}.

\subsection{Investigating Neighbor information through Tuned Lens}
\label{sec:experiments:tunedlens}

To further analyze the latent representations learned by our models, we apply the Tuned Lens method introduced in \cite{belroseElicitingLatentPredictions2023}.

The Tuned Lens provides a direct view into the information encoded at each layer, $l$, by learning a linear transformation $\mathbf{L}_{l}: \mathbb{R}^{d_{\texttt{model}}} \to \mathbb{R}^{d_{\texttt{model}}}$ (referred to as a ``translator'') which attempts to map embeddings to their final state (after the last layer), i.e., $\textbf{L}_{l}(R_l(t)) = R_{l_{\texttt{final}}}(t)$. By applying these learned translators, we are able to unembed (into the vocabulary) embeddings from any layer in the model, thus gaining insight into what the model has captured after performing a few layers of computation.

\begin{figure}
     \centering
     \makebox[\textwidth][c]{
     \begin{subfigure}[b]{0.48\textwidth}
         \centering
         \includegraphics[height=5cm]{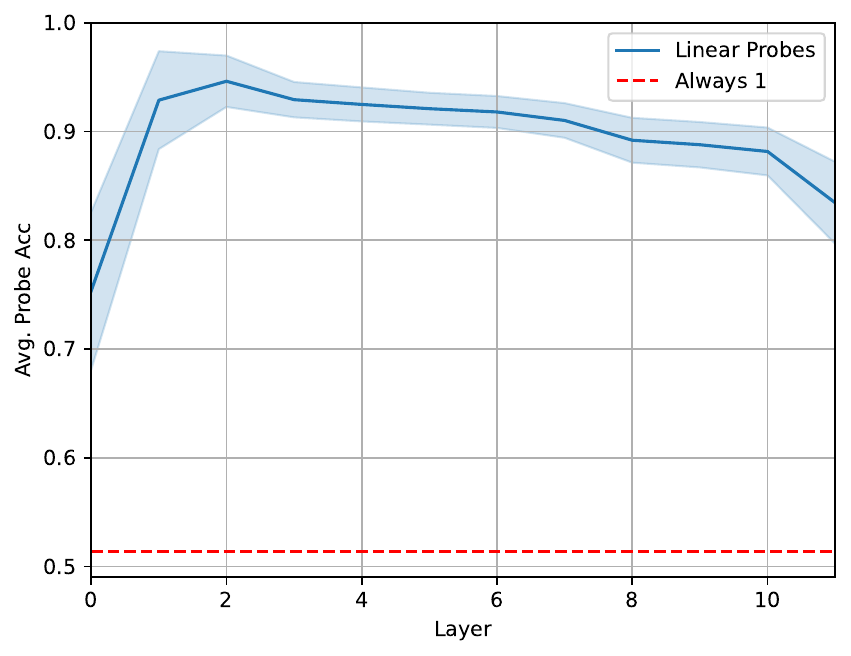}
         \caption{Accuracy of linear probes averaged across all coordinate positions and wall directions in 15,000 mazes. In the appendix, we show that across models with high accuracy and a linearly decodable representation of the maze ($>90\%$ accuracy), $4/5$ of these were most effective at layer 2, and $1/5$ at layer 3. Models that performed poorly often acquired such representations at later layers, but the resulting accuracies of probe sets never exceeded $80\%$.}
         \label{fig:experiments.probes.layerwise}
     \end{subfigure}
     \hfill
     \begin{subfigure}[b]{0.48\textwidth}
         \centering
         \includegraphics[height=5cm]{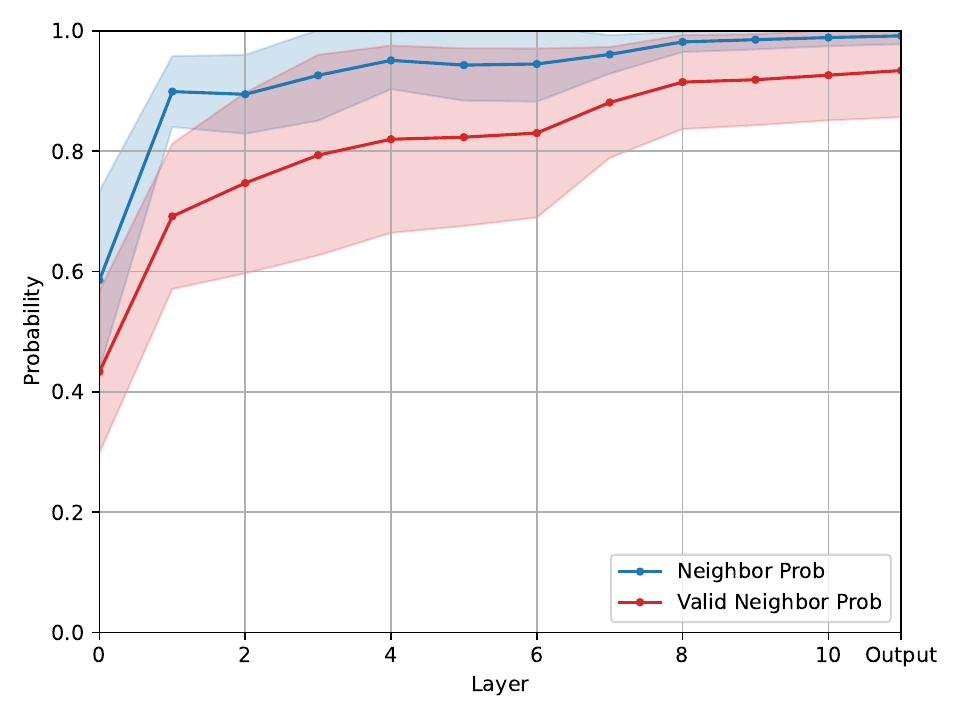}
         \caption{Results from the TunedLens. For each layer, we compute the probability mass given to \textcolor{mplred}{connected}, or \textcolor{mpldarkblue}{merely adjacent} neighboring coordinates for all coordinate tokens in the path gathered across 50 rollouts. We observe that valid neighbors become relatively more probable after layer 2, and the most significant variations in neighbor probability are aligned with the layers in the model when the maze representation is most clear.
         }
         \label{fig:experiments.probing.tunedlens}
     \end{subfigure}
     }
        \caption{Analysis of Linear Probes and Tunes Lens applied to the \texttt{<PATH\_START>} token for the most performant transformer (\jirpy ). Colored regions correspond to $1\sigma$.}
        \label{fig:tunedlens}
\end{figure}

We apply the Tuned Lens approach to see at which layers models write information about neighbors onto coordinate tokens; this includes connected neighbors (those not blocked by walls) and all neighbors (all adjacent coordinates). The resulting analysis for the \jirpy \ model is shown in \autoref{fig:experiments.probing.tunedlens}.
We see that after the first layer, the residual stream already contains significant information concerning whether the next token in a path will be a coordinate, coinciding with the layer in the model where a linear representation of the maze is captured most clearly. This information is then refined gradually throughout the model, such that at later layers, the validity of the next token is enforced more strongly, perhaps owing to the effects of the heads identified in (\autoref{sec:experiments:dla}). 

\subsection{When Do Models Learn to Represent the Maze?}
\label{sec:experiments:grokking}

\begin{figure}
     \centering
     \makebox[\textwidth][c]{
     \begin{subfigure}[c]{0.45\textwidth}
         \centering
         \includegraphics[width=\textwidth]{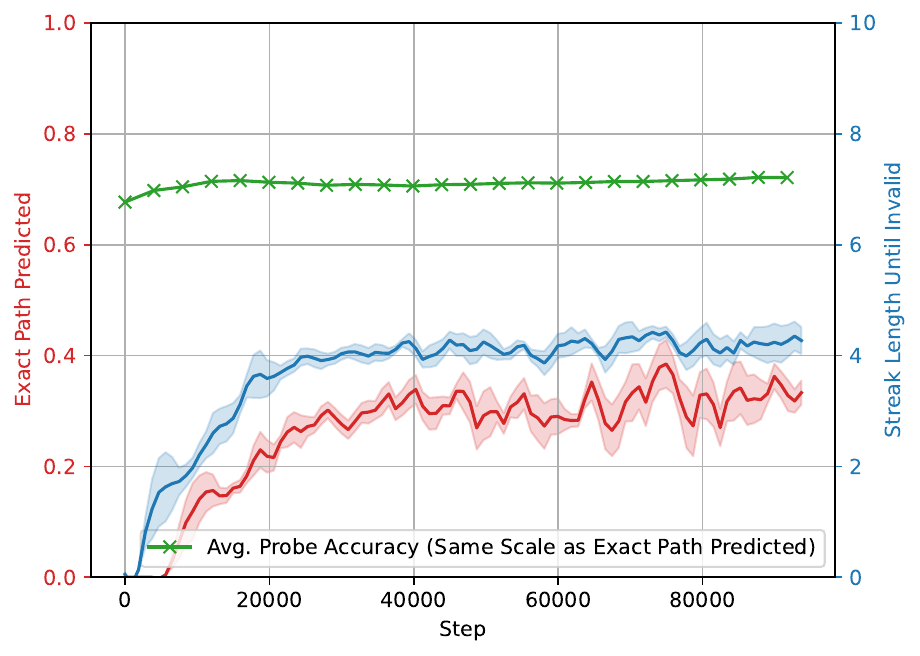}
         \caption{\hallway}
         \label{fig:PATH_START_wm_acc_against_test_acc_hallway}
     \end{subfigure}
     \hfill
     \begin{subfigure}[c]{0.45\textwidth}
         \centering
         \includegraphics[width=\textwidth]{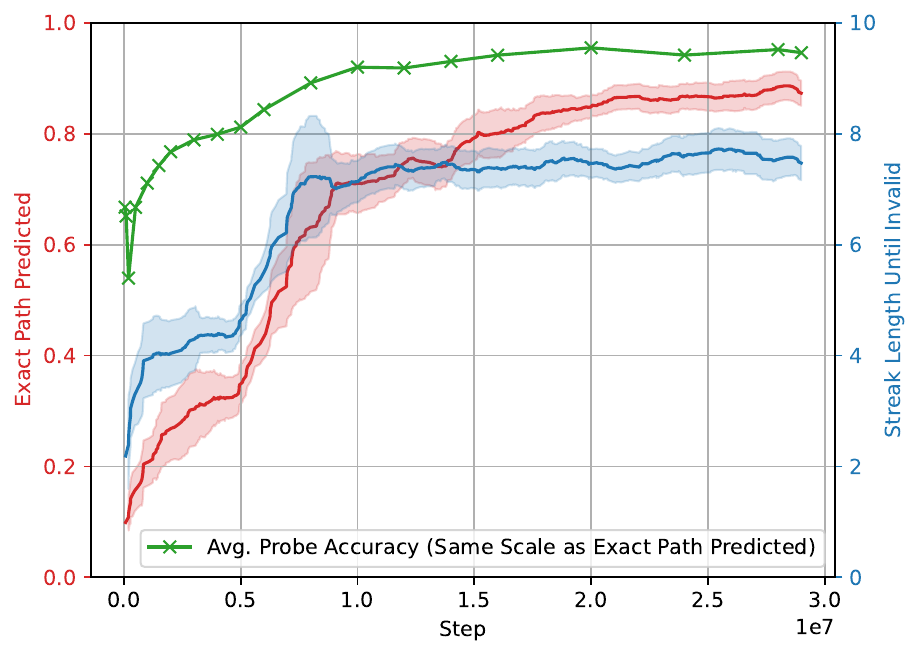}
         \caption{\jirpy}
         \label{fig:PATH_START_wm_acc_against_test_acc_jirpy}
     \end{subfigure}
     }
        \caption{Layerwise analysis of maze structure captured by the model. Note that the distribution of paths for hallway mazes is shorter than those for forking mazes.}
        \label{fig:experiments.training_wms}
\end{figure}

Prior work has shown that the phenomenon of grokking \cite{nandaProgressMeasuresGrokking2023, liuUnderstandingGrokkingEffective2022}, in which the test accuracy (i.e., the generalizability of a model's learned behavior) improves abruptly during training, may be linked to the formation of structured representations over which a task can be robustly solved \cite{liuUnderstandingGrokkingEffective2022}. As we established in \autoref{sec:experiments:worldmodels} that models learn linearly structured representations of mazes, it is a natural question to ask when these are learned and if they co-occur with any notable changes in a model's performance during training. To this end, we trained probe sets across checkpoints for both the \hallway \ and \jirpy \ models, with results shown in \autoref{fig:experiments.training_wms}. We find that the \hallway \ model does not learn a clear linear representation of the maze, while \jirpy \ does. Furthermore, the periods during training in which these representations improve the most also correspond to the times at which the model's performance improves most sharply. This provides suggestive but incomplete evidence for the possibility that these representations play a causal role in the model's behavior.

\newpage
\section{Related Work}

Transformers' ability to solve inherently difficult tasks is increasingly being explored. 
In particular, the capability of transformers to process semantic information and emulate program behavior, especially with structural recursion, has been investigated \cite{zhangCanTransformersLearn2023, momennejadEvaluatingCognitiveMaps2023, liu2023evaluating}. 

Other research on transformers suggests that some performance may be attributed to an architectural bias towards mesa-optimization \cite{vonoswaldUncoveringMesaoptimizationAlgorithms2023}. Here, it is argued that transformers employ mesa-optimization during their forward pass, constructing an internal learning objective and optimizing it. Akyurek et al. note that transformers might harness standard learning algorithms implicitly, encoding miniature models within their activations and updating these based on incoming examples~\cite{akyurek2023learning}.

\textbf{Finding Meaningful Directions in Activation Space}:
Mechanistic interpretability seeks to reverse engineer neural networks. In the pursuit of this ambitious approach to interpretability, several techniques have been proposed in an effort to find and understand meaningful directions in a model's activation spaces. Belrose et al.'s Tuned Lens \cite{belroseElicitingLatentPredictions2023} involves training affine transformations that translate the basis associated with representations in any single layer's activation space with the expected basis of that of the final layer. Such transformations, when coupled with the model's unembedding layer, can be used to map the residual stream to a distribution over the model's vocabulary. 

Sparse Coding employs autoencoders augmented with sparsity regularization to derive disentangled representations of an activation space; this approach has been researched in works by Bricken et al. \cite{bricken2023monosemanticity}. Other efforts to find meaningful directions in activation space include using $k$-sparse linear classifiers that map the activations of a single neuron or a collection of neurons to specific features \cite{gurneeFindingNeuronsHaystack2023}.

\textbf{World Models}:

Much recent research has been focused around finding and understanding world models, especially in planning tasks. 
Li et al. \cite{liEmergentWorldRepresentations2023} studied world representations in the game of Othello, with Nanda et al. \cite{nanda_othello_2023} further investigating the linearity of these representations. 
Turner et al. \cite{turntroutUnderstandingControllingMazesolving} focused on reinforcement learning, examining maze-solving tasks and the underlying representations. 
Additionally, the introduction of mechanistic interpretability for decision transformers by Bloom et al. \cite{bloomMechanisticInterpretabilityAnalysis} offers a new perspective on interpretability in strategic planning tasks. 
Together, these studies provide valuable insights into the structure and utility of world models across different contexts.

\vspace{2em}

\section{Conclusion}

We demonstrate that transformers trained to solve mazes acquire emergent linear representations that capture maze connectivity and are encoded in a single token's latent state. The embedding layer of trained models is shown to learn an emergent spatial structure. Furthermore, we find that in simple models, some attention heads learn to respect the topology of the maze and present some evidence and hypotheses as to their function. In future work, we aim to construct a more complete mechanistic picture of how these elementary heads operate over the linear world model and ultimately form circuits responsible for solving mazes. Additionally, future work will investigate the generality of such emergent models by investigating distinct classes of neural networks trained to perform the same tasks over entirely different input representations in an attempt to provide further evidence for claims of representational ``universality''.
With this work, we hope to inspire other researchers to investigate the seemingly systematic yet elusive internal behavior of transformer models.

\vspace{2em}

\begin{ack}
We are grateful to AI Safety Camp for initially supporting this project and bringing many of the authors together.
This work was partially funded by National Science Foundation awards DMS-2110745 and DMS-2309810. 
\end{ack}

\newpage
{
\footnotesize
\bibliography{references}
}

\newpage

\appendix
A complete set of notebooks and supplementary data can be found with our codebase: \\
\href{https://github.com/understanding-search/structured-representations-maze-transformers}{\texttt{github.com/understanding-search/structured-representations-maze-transformers}}

\section{Model Hyperparameters}
\label{overveiw}
\begin{table*}
    \centering
    \begin{center}
    \begin{small}

    \begin{tabular}{lll}
    \dtoprule
    Model & Component(s) & Details \\ \cmidrule(r){1-1} \cmidrule(lr){2-2} \cmidrule(l){3-3} 
    \multirow{2}{*}{Jirpy} & Transformer Decoder & $32$  \\
                           & Optimizer & adamW, lr=0.0023 \vspace{0.2cm} \\ 
    \dbottomrule
    \end{tabular}
    
    \end{small}
    \end{center}
    \vskip -0.1in
    \caption{Architectural Details} 
    \label{tab:architectural_details}
\end{table*}

\begin{table}
    \footnotesize
    \centering
    \begin{tabular}{lrrrccc}
        \toprule
        Model &  $d_{\texttt{model}}$ &  $d_{\texttt{head}}$ &  $n_{\texttt{layers}}$ & total parameters & dataset config & training dataset size \\ 
        \midrule
        \hallway &      128 &      32 &         6 & $\approx 9.64 \cdot 10^6$ & \texttt{dfs(do\_forks=false)} & $3 \times 10^{6}$  \\
        \jirpy   &      256 &      16 &        12 & $\approx 1.24 \cdot 10^6$ & varied                        &$5 \times 10^{6}$  (6 Epochs) \\
        \bottomrule
    \end{tabular}
    \setlength{\abovecaptionskip}{8pt}
    \caption{Hyperparameters for the two models we investigate. Both models are trained via the \texttt{AdamW} optimizer with a learning rate of $10^{-4}$ and batch size of 32. A single training epoch is performed.}
    \label{tab:hparams}
\end{table}

\section{Performance Statistics}
\label{sec:performance-APPENDIX}

\begin{table}
    \centering
    { \footnotesize 
        \begin{tabular}{l|rrrrrr}
\toprule
               \multicolumn{1}{r}{dataset:} & \multicolumn{2}{c}{forkless} & \multicolumn{2}{c}{RDFS} & \multicolumn{2}{c}{pRDFS} \\
                 \multicolumn{1}{r}{model:} & \hallway &   \jirpy & \hallway &    \jirpy &  \hallway &   \jirpy \\
\midrule
\midrule
                  exactly correct rollouts &  36.3\% &   52.7\% &   36.7\% &    29.3\% &    17.6\% &  21.1\% \\
                            valid rollouts &  50.8\% &   59.8\% &   41.4\% &    29.7\% &    74.2\% &  26.6\% \\
              rollouts with target reached &  87.1\% &   80.5\% &   76.6\% &    98.4\% &    28.9\% &  98.4\% \\
\midrule
                      \texttt{path\_start} & 100.0\% &  100.0\% &  100.0\% &   100.0\% &   100.0\% & 100.0\% \\
       \texttt{origin\_after\_path\_start} &  94.5\% &   95.3\% &  100.0\% &   100.0\% &    97.3\% & 100.0\% \\
              \texttt{first\_path\_choice} &  70.3\% &   70.3\% &   65.6\% &    60.2\% &    66.8\% &  54.3\% \\
                \texttt{rand\_path\_token} &  91.0\% &   89.8\% &   91.8\% &    88.3\% &    86.7\% & 78.1\% \\
 \texttt{rand\_path\_token\_nonend} &  96.9\% &   93.8\% &   97.3\% &    88.7\% &    83.6\% & 79.7\% \\
         \texttt{final\_before\_path\_end} &  98.8\% &   87.5\% &   94.5\% &    95.7\% &    82.8\% & 90.6\% \\
                        \texttt{path\_end} &  88.3\% &   81.6\% &  100.0\% &   100.0\% &    99.6\% & 100.0\% \\
\bottomrule
\end{tabular}
    }
    \vspace{0.8em}
    \caption{Same as \autoref{tab:task-accuracies}, but for $7 \times 7$ forkless, pRDFS, and RDFS mazes (See \autoref{sec:dataset} and \cite{ivanitskiyConfigurableLibraryGenerating2023}), also with $n=256$ samples. The second group of single-token tasks used to assess performance are detailed in \autoref{fig:task-desc}. Note that the \jirpy \ model was trained only mazes that were only dense in $6 \times 6$ subgrids, and generalizes relatively poorly to these larger mazes. Conversely, the \hallway \ model was trained on only sparse mazes to start with, and performs similarly.}
    \label{tab:task-accuracies-g7-APPENDIX}
\end{table}

\begin{table}
    \centering
    { \footnotesize 
        \begin{tabular}{l|rrrrrr}
\toprule
             \multicolumn{1}{r}{dataset:}   & \multicolumn{2}{c}{forkless} & \multicolumn{2}{c}{RDFS} & \multicolumn{2}{c}{pRDFS} \\
             \multicolumn{1}{r}{model:}     & \hallway &   \jirpy & \hallway &    \jirpy &  \hallway &   \jirpy \\
\midrule
\midrule
                  exactly correct rollouts &  \textbf{38.3\%} &   \textbf{38.7\%} &   24.2\% &    82.4\% &    24.2\% &    70.7\% \\
                            valid rollouts &  \textbf{54.3\%} &   \textbf{53.5\%} &   37.5\% &    84.0\% &    49.6\% &    87.1\% \\
              rollouts with target reached &  \textbf{87.1\%} &   \textbf{64.5\%} &   94.5\% &    99.2\% &    92.6\% &   100.0\% \\
\midrule
                     \texttt{path\_start}  & 100.0\% &  100.0\% &  100.0\% &   100.0\% &   100.0\% &   100.0\% \\
       \texttt{origin\_after\_path\_start} &  91.0\% &   86.7\% &  100.0\% &   100.0\% &   100.0\% &   100.0\% \\
              \texttt{first\_path\_choice} &  \textbf{71.5\%} &   \textbf{66.4\%} &   \textbf{67.2\%} &    \textbf{86.7\%} &    \textbf{66.4\%} &    \textbf{84.4\%} \\
                \texttt{rand\_path\_token} &  93.0\% &   87.1\% &   90.2\% &    98.0\% &    84.0\% &    94.5\% \\
 \texttt{rand\_path\_token\_nonend} &  97.3\% &   89.8\% &   92.2\% &    99.2\% &    84.4\% &    97.3\% \\
         \texttt{final\_before\_path\_end} &  95.7\% &   85.9\% &   93.4\% &   100.0\% &    84.4\% &   100.0\% \\
                        \texttt{path\_end} &  86.7\% &   71.5\% &  100.0\% &    99.6\% &   100.0\% &   100.0\% \\
\bottomrule
\end{tabular}

    }
    \vspace{0.8em}
    \caption{An exact repeat of \autoref{tab:task-accuracies}, provided here for ease of comparison with \autoref{tab:task-accuracies-g7-APPENDIX}.}
    \label{tab:task-accuracies-g6-APPENDIX}
\end{table}

\begin{figure}
     \centering
     \makebox[\textwidth][c]{
     \begin{subfigure}[b]{0.38\textwidth}
        \centering
        \includegraphics[width=1\textwidth]{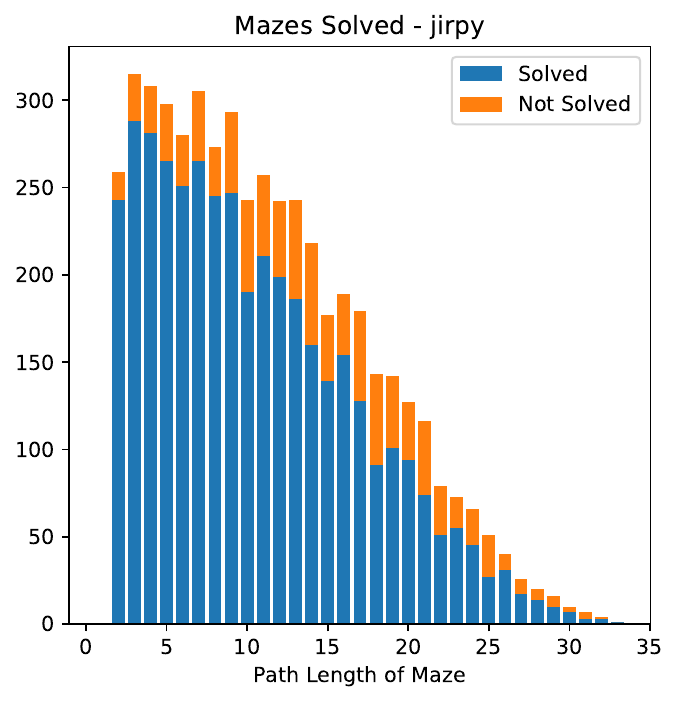}
        \caption{\jirpy }
     \end{subfigure}
     
     \begin{subfigure}[b]{0.38\textwidth}
         \centering
          \includegraphics[width=1\textwidth]{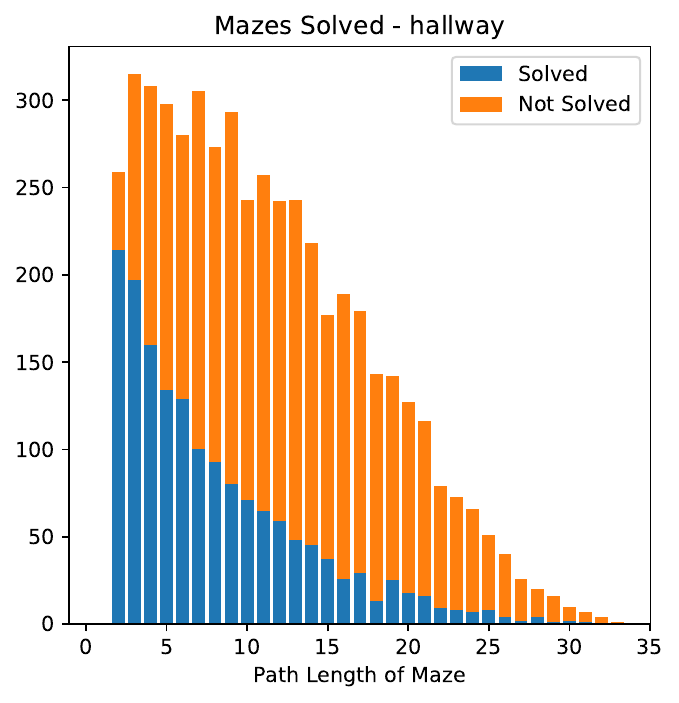}
        \caption{\hallway }
     \end{subfigure}
     
      \begin{subfigure}[b]{0.38\textwidth}
         \centering
          \includegraphics[width=1\textwidth]{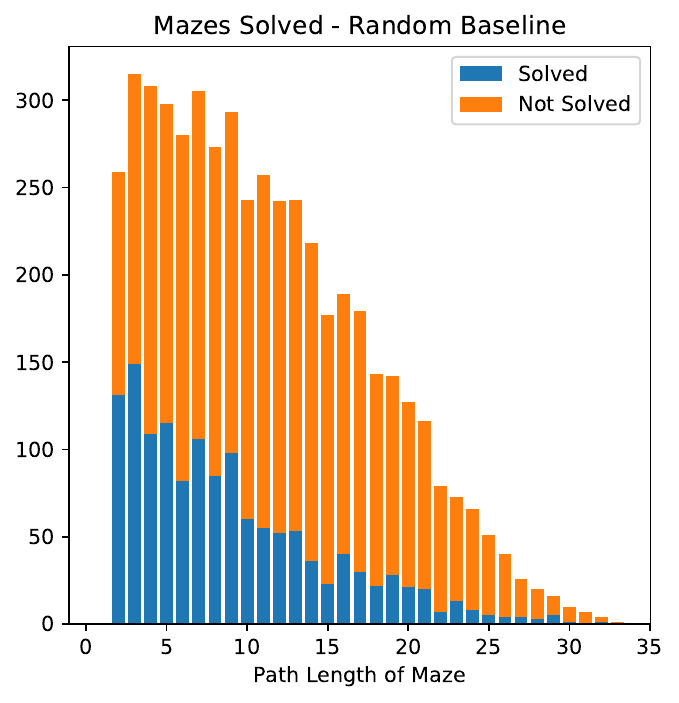}
        \caption{Random Baseline}
     \end{subfigure}
     }
    \caption{Performance of our models on a held-out test set of RDFS mazes. Noticeably, the performance appears roughly consistent over path length for \jirpy \ which was trained to solve mazes, and similarly for the random baseline (which follows corridors and chooses a random continuation of its path when reaching a fork). The hallway model, on the other hand, is able to ``solve" some of the very short mazes but struggles with longer mazes (where it is likely to encounter forking points).}
    \label{fig:three graphs}
\end{figure}

\begin{figure}
    \centering
    \includegraphics[width=0.9\textwidth]{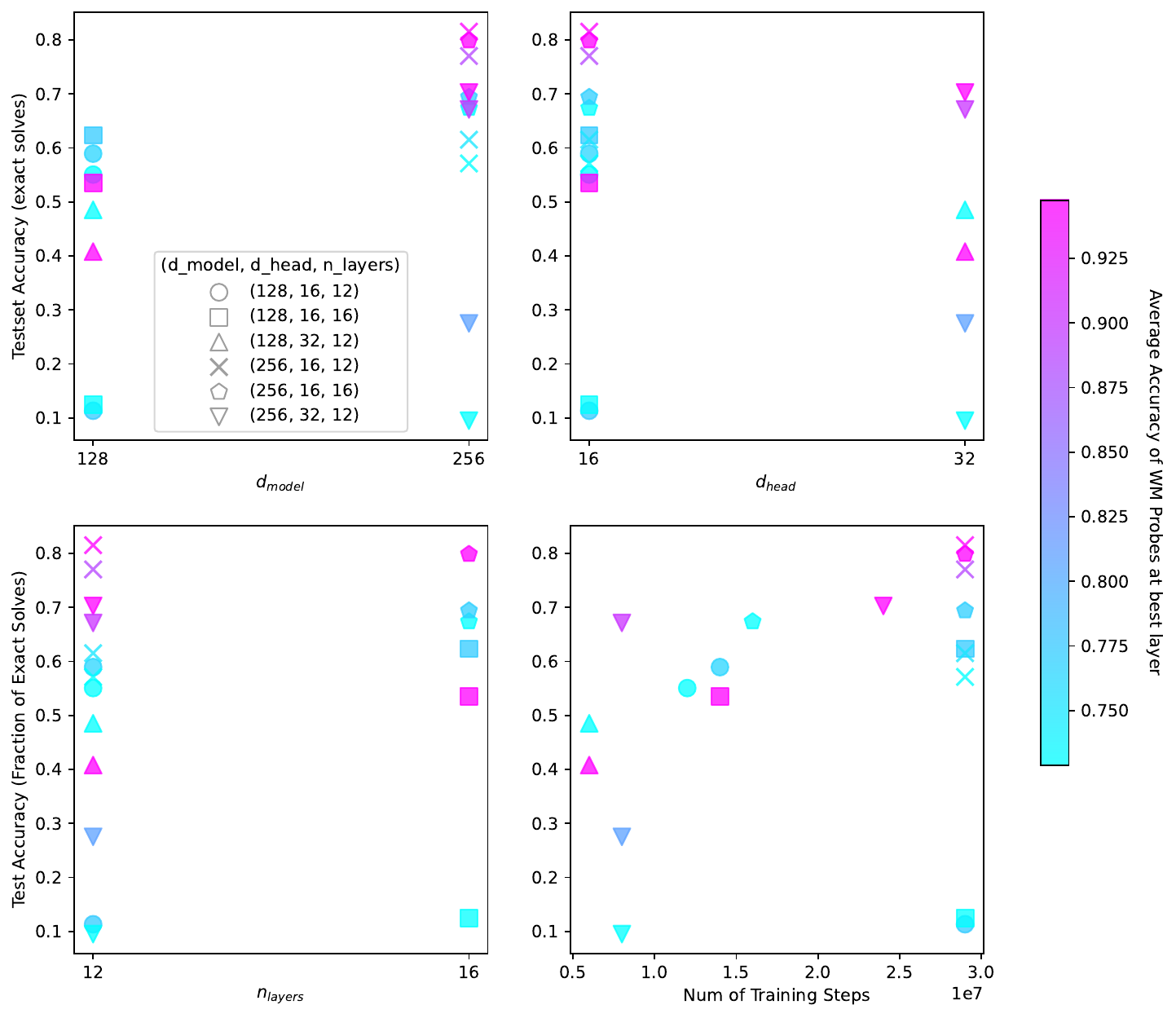}
    \caption{Results of sweeps carried out across a variety of models which were trained for varying lengths. We find that 1) Models can do well even if they don’t have a linear maze representation, which we can decode, but the very best models also have a linear maze representation. 2) $d_{model}$ and training time are the only hyperparameters that seemed to correlate with performance in the regimes we considered, but any such correlations are too weak to draw strong conclusions. The model with the highest test accuracy and probe accuracy is \jirpy.}
    \label{fig:app:sweep_performance}
\end{figure}

\newpage
\section{Embedding Structure}
\label{sec:appendix:embed}

Appendix of embedding structure results to \autoref{fig:embed-structure}.

\begin{figure}[ht]
    \centering
    \begin{minipage}[b]{0.45\textwidth}
        \includegraphics[width=\textwidth]{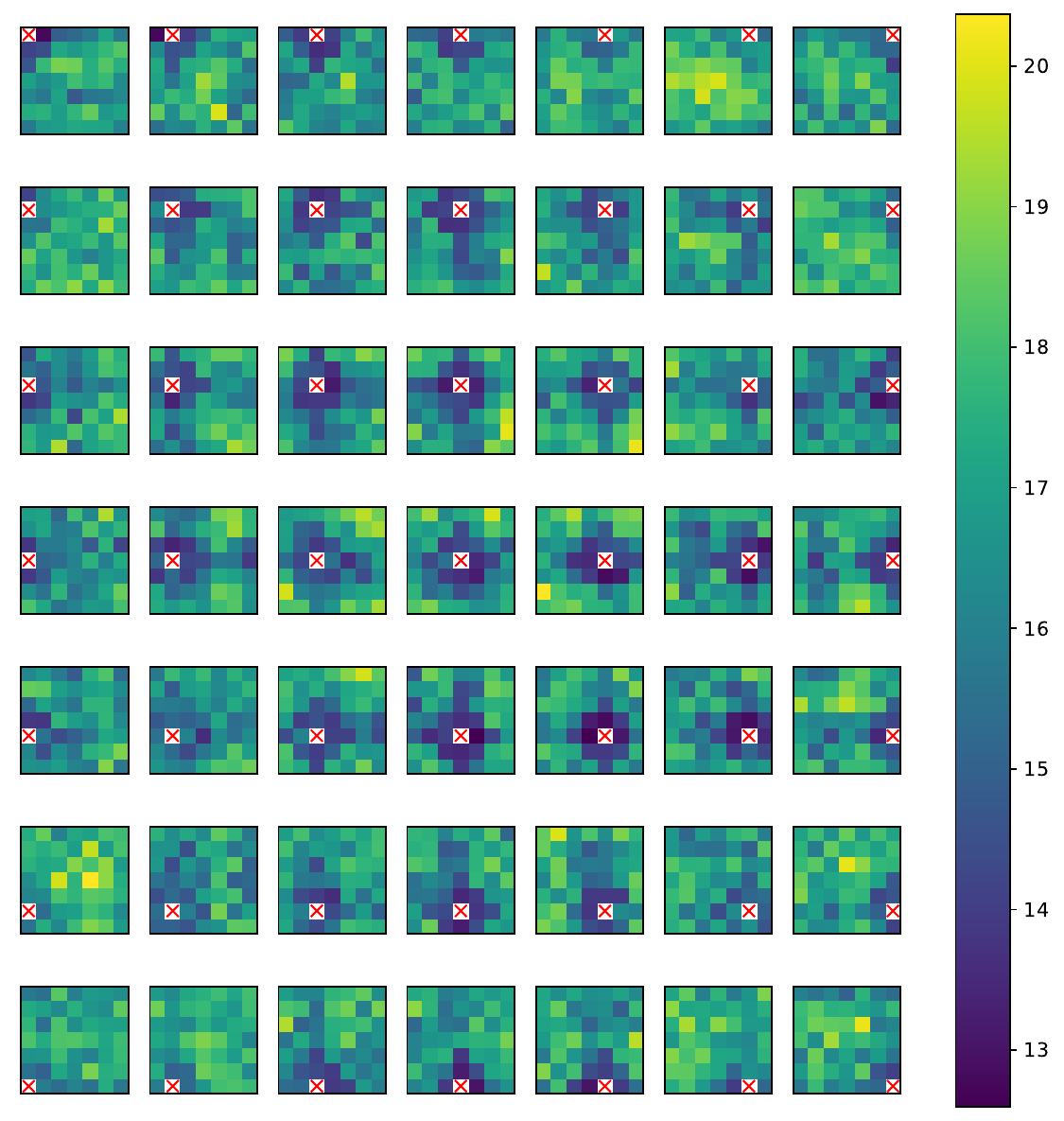}
        \caption{Given the coordinate at the \textcolor{red}{\texttt{x}}, we view the Manhattan distance to the other tokens on the grid for the \hallway \ model.}
        \label{fig:hallway-distgrid-APPENDIX}
    \end{minipage}
    \hfill
    \begin{minipage}[b]{0.45\textwidth}
        \includegraphics[width=\textwidth]{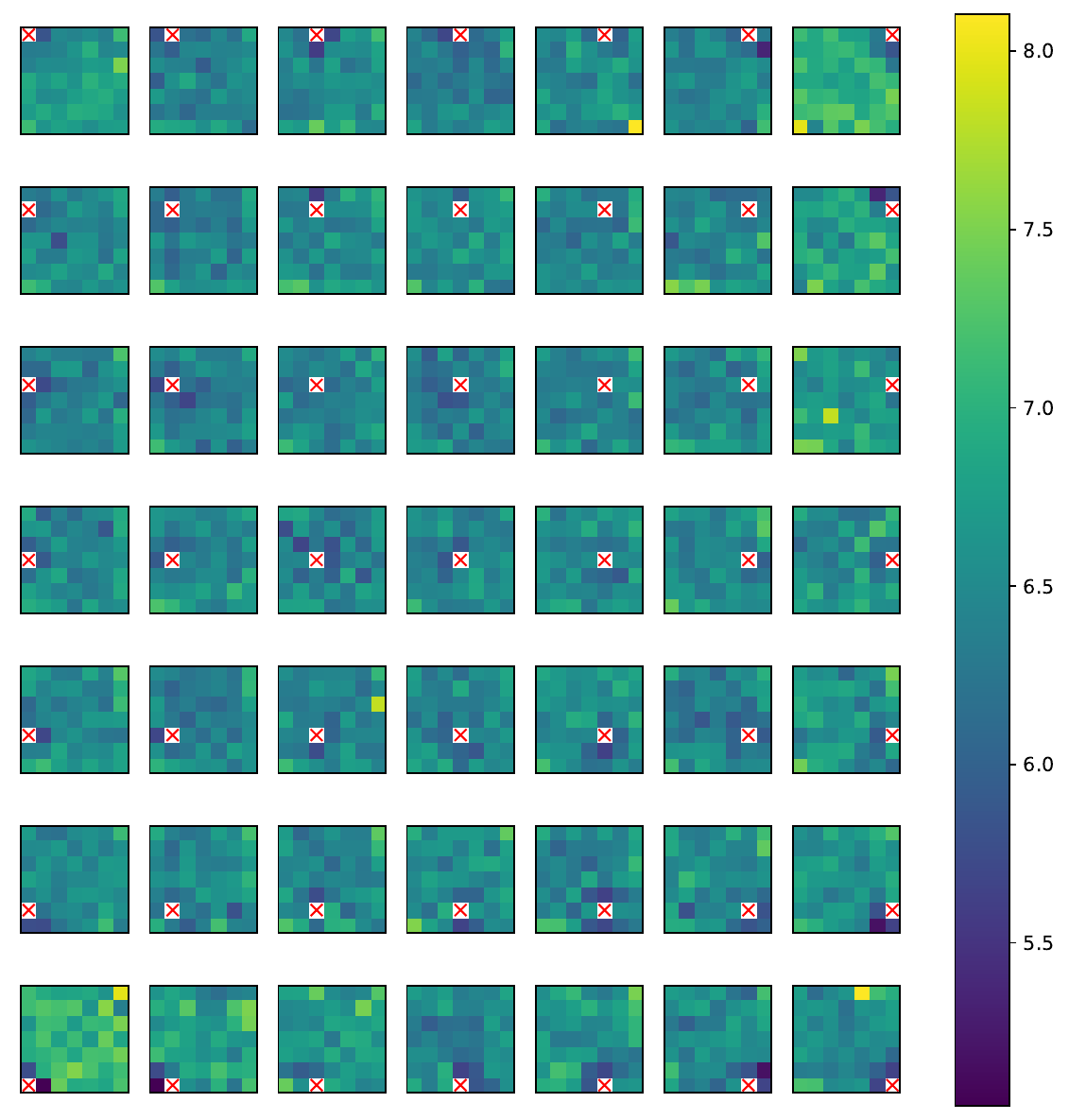}
        \caption{Given the coordinate at the \textcolor{red}{\texttt{x}}, we view the Manhattan distance to the other tokens on the grid for the \jirpy \ model.}
        \label{fig:jirpy-distgrid-APPENDIX}
    \end{minipage}
\end{figure}

\begin{figure}
    \centering
    \begin{minipage}[t]{0.5\textwidth}
\begin{tikzpicture}

\definecolor{darkgray176}{RGB}{176,176,176}
\definecolor{darkslategray63}{RGB}{63,63,63}
\definecolor{steelblue49115161}{RGB}{49,115,161}

\begin{axis}[
tick align=outside,
tick pos=left,
x grid style={darkgray176},
xlabel={$\Vert a - b \Vert_1$},
xmin=-0.5, xmax=11.5,
xtick style={color=black},
xtick={0,1,2,3,4,5,6,7,8,9,10,11},
xticklabels={1.0,2.0,3.0,4.0,5.0,6.0,7.0,8.0,9.0,10.0,11.0,12.0},
y grid style={darkgray176},
ylabel={$\Vert E(a) - E(b) \Vert_1$},
ymin=5.0, ymax=8.5,
ytick style={color=black}
]
\path [draw=darkslategray63, fill=steelblue49115161, semithick]
(axis cs:-0.25,5.94471519540093)
--(axis cs:0.25,5.94471519540093)
--(axis cs:0.25,6.39002881958731)
--(axis cs:-0.25,6.39002881958731)
--(axis cs:-0.25,5.94471519540093)
--cycle;
\path [draw=darkslategray63, fill=steelblue49115161, semithick]
(axis cs:0.75,6.16578658808066)
--(axis cs:1.25,6.16578658808066)
--(axis cs:1.25,6.50461844100937)
--(axis cs:0.75,6.50461844100937)
--(axis cs:0.75,6.16578658808066)
--cycle;
\path [draw=darkslategray63, fill=steelblue49115161, semithick]
(axis cs:1.75,6.45318817748193)
--(axis cs:2.25,6.45318817748193)
--(axis cs:2.25,6.76353628682773)
--(axis cs:1.75,6.76353628682773)
--(axis cs:1.75,6.45318817748193)
--cycle;
\path [draw=darkslategray63, fill=steelblue49115161, semithick]
(axis cs:2.75,6.2821569327516)
--(axis cs:3.25,6.2821569327516)
--(axis cs:3.25,6.62928186239151)
--(axis cs:2.75,6.62928186239151)
--(axis cs:2.75,6.2821569327516)
--cycle;
\path [draw=darkslategray63, fill=steelblue49115161, semithick]
(axis cs:3.75,6.39035022644384)
--(axis cs:4.25,6.39035022644384)
--(axis cs:4.25,6.71056780196159)
--(axis cs:3.75,6.71056780196159)
--(axis cs:3.75,6.39035022644384)
--cycle;
\path [draw=darkslategray63, fill=steelblue49115161, semithick]
(axis cs:4.75,6.35897064908204)
--(axis cs:5.25,6.35897064908204)
--(axis cs:5.25,6.80297659384087)
--(axis cs:4.75,6.80297659384087)
--(axis cs:4.75,6.35897064908204)
--cycle;
\path [draw=darkslategray63, fill=steelblue49115161, semithick]
(axis cs:5.75,6.36248638206598)
--(axis cs:6.25,6.36248638206598)
--(axis cs:6.25,6.73088137192099)
--(axis cs:5.75,6.73088137192099)
--(axis cs:5.75,6.36248638206598)
--cycle;
\path [draw=darkslategray63, fill=steelblue49115161, semithick]
(axis cs:6.75,6.4790466926861)
--(axis cs:7.25,6.4790466926861)
--(axis cs:7.25,6.95138973009307)
--(axis cs:6.75,6.95138973009307)
--(axis cs:6.75,6.4790466926861)
--cycle;
\path [draw=darkslategray63, fill=steelblue49115161, semithick]
(axis cs:7.75,6.51662072139152)
--(axis cs:8.25,6.51662072139152)
--(axis cs:8.25,6.9020416617459)
--(axis cs:7.75,6.9020416617459)
--(axis cs:7.75,6.51662072139152)
--cycle;
\path [draw=darkslategray63, fill=steelblue49115161, semithick]
(axis cs:8.75,6.49244233695572)
--(axis cs:9.25,6.49244233695572)
--(axis cs:9.25,7.22115067957566)
--(axis cs:8.75,7.22115067957566)
--(axis cs:8.75,6.49244233695572)
--cycle;
\path [draw=darkslategray63, fill=steelblue49115161, semithick]
(axis cs:9.75,6.41034861261869)
--(axis cs:10.25,6.41034861261869)
--(axis cs:10.25,6.64540235041204)
--(axis cs:9.75,6.64540235041204)
--(axis cs:9.75,6.41034861261869)
--cycle;
\path [draw=darkslategray63, fill=steelblue49115161, semithick]
(axis cs:10.75,7.06223889967805)
--(axis cs:11.25,7.06223889967805)
--(axis cs:11.25,7.67514804028178)
--(axis cs:10.75,7.67514804028178)
--(axis cs:10.75,7.06223889967805)
--cycle;
\addplot [semithick, darkslategray63]
table {%
0 5.94471519540093
0 5.58232606077217
};
\addplot [semithick, darkslategray63]
table {%
0 6.39002881958731
0 6.94780435040593
};
\addplot [semithick, darkslategray63]
table {%
-0.125 5.58232606077217
0.125 5.58232606077217
};
\addplot [semithick, darkslategray63]
table {%
-0.125 6.94780435040593
0.125 6.94780435040593
};
\addplot [semithick, darkslategray63]
table {%
1 6.16578658808066
1 5.68485272178077
};
\addplot [semithick, darkslategray63]
table {%
1 6.50461844100937
1 7.00761016376055
};
\addplot [semithick, darkslategray63]
table {%
0.875 5.68485272178077
1.125 5.68485272178077
};
\addplot [semithick, darkslategray63]
table {%
0.875 7.00761016376055
1.125 7.00761016376055
};
\addplot [semithick, darkslategray63]
table {%
2 6.45318817748193
2 6.03409436163201
};
\addplot [semithick, darkslategray63]
table {%
2 6.76353628682773
2 7.17186426313128
};
\addplot [semithick, darkslategray63]
table {%
1.875 6.03409436163201
2.125 6.03409436163201
};
\addplot [semithick, darkslategray63]
table {%
1.875 7.17186426313128
2.125 7.17186426313128
};
\addplot [semithick, darkslategray63]
table {%
3 6.2821569327516
3 5.77628982477472
};
\addplot [semithick, darkslategray63]
table {%
3 6.62928186239151
3 7.14255523023894
};
\addplot [semithick, darkslategray63]
table {%
2.875 5.77628982477472
3.125 5.77628982477472
};
\addplot [semithick, darkslategray63]
table {%
2.875 7.14255523023894
3.125 7.14255523023894
};
\addplot [semithick, darkslategray63]
table {%
4 6.39035022644384
4 6.07195477110508
};
\addplot [semithick, darkslategray63]
table {%
4 6.71056780196159
4 7.15641726268223
};
\addplot [semithick, darkslategray63]
table {%
3.875 6.07195477110508
4.125 6.07195477110508
};
\addplot [semithick, darkslategray63]
table {%
3.875 7.15641726268223
4.125 7.15641726268223
};
\addplot [semithick, darkslategray63]
table {%
5 6.35897064908204
5 5.96004400754464
};
\addplot [semithick, darkslategray63]
table {%
5 6.80297659384087
5 7.39281084154209
};
\addplot [semithick, darkslategray63]
table {%
4.875 5.96004400754464
5.125 5.96004400754464
};
\addplot [semithick, darkslategray63]
table {%
4.875 7.39281084154209
5.125 7.39281084154209
};
\addplot [semithick, darkslategray63]
table {%
6 6.36248638206598
6 5.87272318624309
};
\addplot [semithick, darkslategray63]
table {%
6 6.73088137192099
6 7.20176055005868
};
\addplot [semithick, darkslategray63]
table {%
5.875 5.87272318624309
6.125 5.87272318624309
};
\addplot [semithick, darkslategray63]
table {%
5.875 7.20176055005868
6.125 7.20176055005868
};
\addplot [semithick, darkslategray63]
table {%
7 6.4790466926861
7 5.9971593428927
};
\addplot [semithick, darkslategray63]
table {%
7 6.95138973009307
7 7.51149008772336
};
\addplot [semithick, darkslategray63]
table {%
6.875 5.9971593428927
7.125 5.9971593428927
};
\addplot [semithick, darkslategray63]
table {%
6.875 7.51149008772336
7.125 7.51149008772336
};
\addplot [semithick, darkslategray63]
table {%
8 6.51662072139152
8 6.15909959198325
};
\addplot [semithick, darkslategray63]
table {%
8 6.9020416617459
8 7.46208564843982
};
\addplot [semithick, darkslategray63]
table {%
7.875 6.15909959198325
8.125 6.15909959198325
};
\addplot [semithick, darkslategray63]
table {%
7.875 7.46208564843982
8.125 7.46208564843982
};
\addplot [semithick, darkslategray63]
table {%
9 6.49244233695572
9 6.3654000958486
};
\addplot [semithick, darkslategray63]
table {%
9 7.22115067957566
9 7.55481835384853
};
\addplot [semithick, darkslategray63]
table {%
8.875 6.3654000958486
9.125 6.3654000958486
};
\addplot [semithick, darkslategray63]
table {%
8.875 7.55481835384853
9.125 7.55481835384853
};
\addplot [semithick, darkslategray63]
table {%
10 6.41034861261869
10 6.14066341120633
};
\addplot [semithick, darkslategray63]
table {%
10 6.64540235041204
10 6.73551877419231
};
\addplot [semithick, darkslategray63]
table {%
9.875 6.14066341120633
10.125 6.14066341120633
};
\addplot [semithick, darkslategray63]
table {%
9.875 6.73551877419231
10.125 6.73551877419231
};
\addplot [semithick, darkslategray63]
table {%
11 7.06223889967805
11 6.75578432937618
};
\addplot [semithick, darkslategray63]
table {%
11 7.67514804028178
11 7.98160261058365
};
\addplot [semithick, darkslategray63]
table {%
10.875 6.75578432937618
11.125 6.75578432937618
};
\addplot [semithick, darkslategray63]
table {%
10.875 7.98160261058365
11.125 7.98160261058365
};
\addplot [semithick, darkslategray63]
table {%
-0.25 6.17524401115952
0.25 6.17524401115952
};
\addplot [semithick, darkslategray63]
table {%
0.75 6.31590530178801
1.25 6.31590530178801
};
\addplot [semithick, darkslategray63]
table {%
1.75 6.60199860521243
2.25 6.60199860521243
};
\addplot [semithick, darkslategray63]
table {%
2.75 6.44703647319693
3.25 6.44703647319693
};
\addplot [semithick, darkslategray63]
table {%
3.75 6.5180361617895
4.25 6.5180361617895
};
\addplot [semithick, darkslategray63]
table {%
4.75 6.58684824256852
5.25 6.58684824256852
};
\addplot [semithick, darkslategray63]
table {%
5.75 6.58551535848528
6.25 6.58551535848528
};
\addplot [semithick, darkslategray63]
table {%
6.75 6.70391086823656
7.25 6.70391086823656
};
\addplot [semithick, darkslategray63]
table {%
7.75 6.7486958455047
8.25 6.7486958455047
};
\addplot [semithick, darkslategray63]
table {%
8.75 6.69444833478337
9.25 6.69444833478337
};
\addplot [semithick, darkslategray63]
table {%
9.75 6.59587152019958
10.25 6.59587152019958
};
\addplot [semithick, darkslategray63]
table {%
10.75 7.36869346997992
11.25 7.36869346997992
};
\end{axis}

\end{tikzpicture}
    \end{minipage}
    \caption{Correlation of coordinate and embedding distance for the \jirpy \ model.}
    \label{fig:jirpy-embed-dist-APPENDIX}
\end{figure}
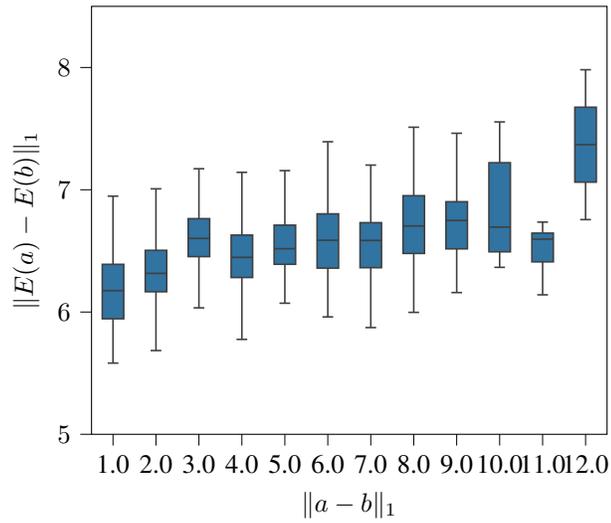

\section{Additional Probing Results}

\begin{figure}
    \centering
    \includegraphics[width=\textwidth]{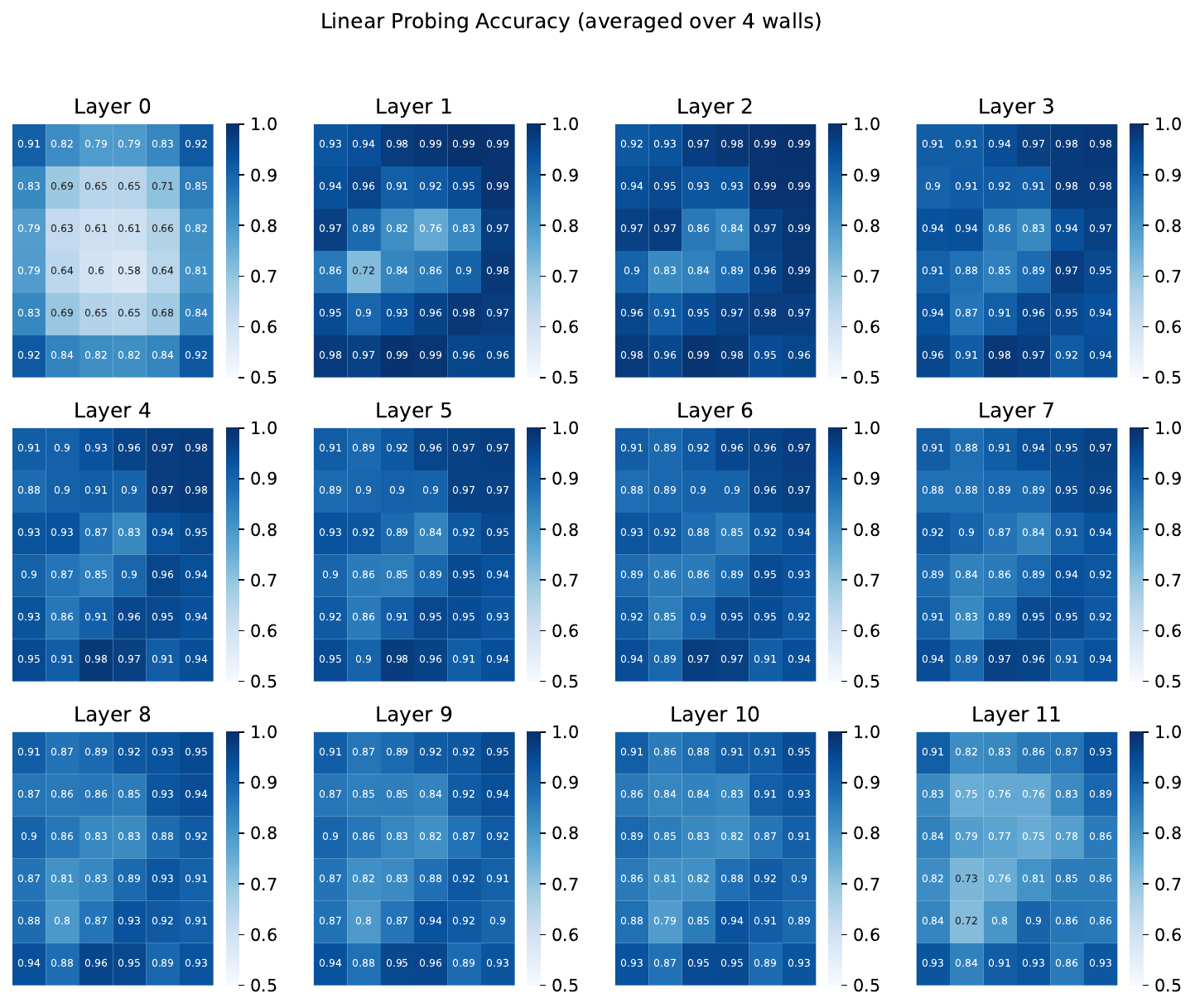}
    \caption{Accuracy of linear probes trained on the residual stream at all layers of the \jirpy \ model. We see that on Layer 2 the highest average probing accuracy is achieved, and the accuracy decreases with later layers. Note that the edges have very high accuracy as outer walls are always present, meaning that 1/4 of the probes on edges (and 2/4 on corners) will achieve 100\% accuracy by always predicting a wall. }
    \label{fig:app:linear_probing_all}
\end{figure}

\begin{figure}
    \centering
    \includegraphics[width=0.7\textwidth]{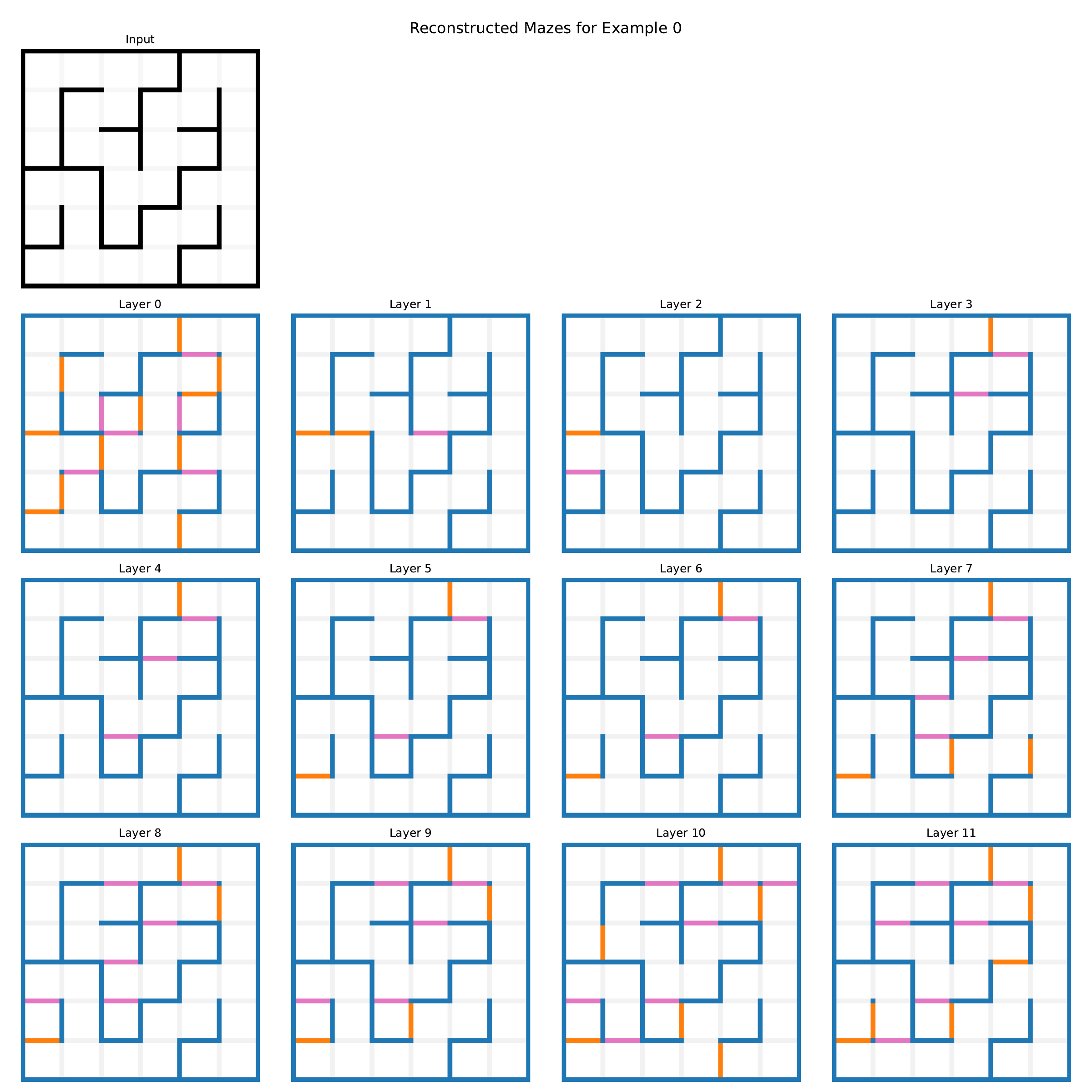}
    \caption{Decoding a maze with probe sets trained for each layer of the \jirpy\ model. Wall colors indicate that thresholded probes \textcolor{mpldarkblue}{Correctly Predicted}, \textcolor{mplorange}{Omitted} or \textcolor{mplpink}{Added} a wall.}
    \label{fig:app:reco_mazes_more_1}
\end{figure}
\begin{figure}
    \centering
    \includegraphics[width=0.7\textwidth]{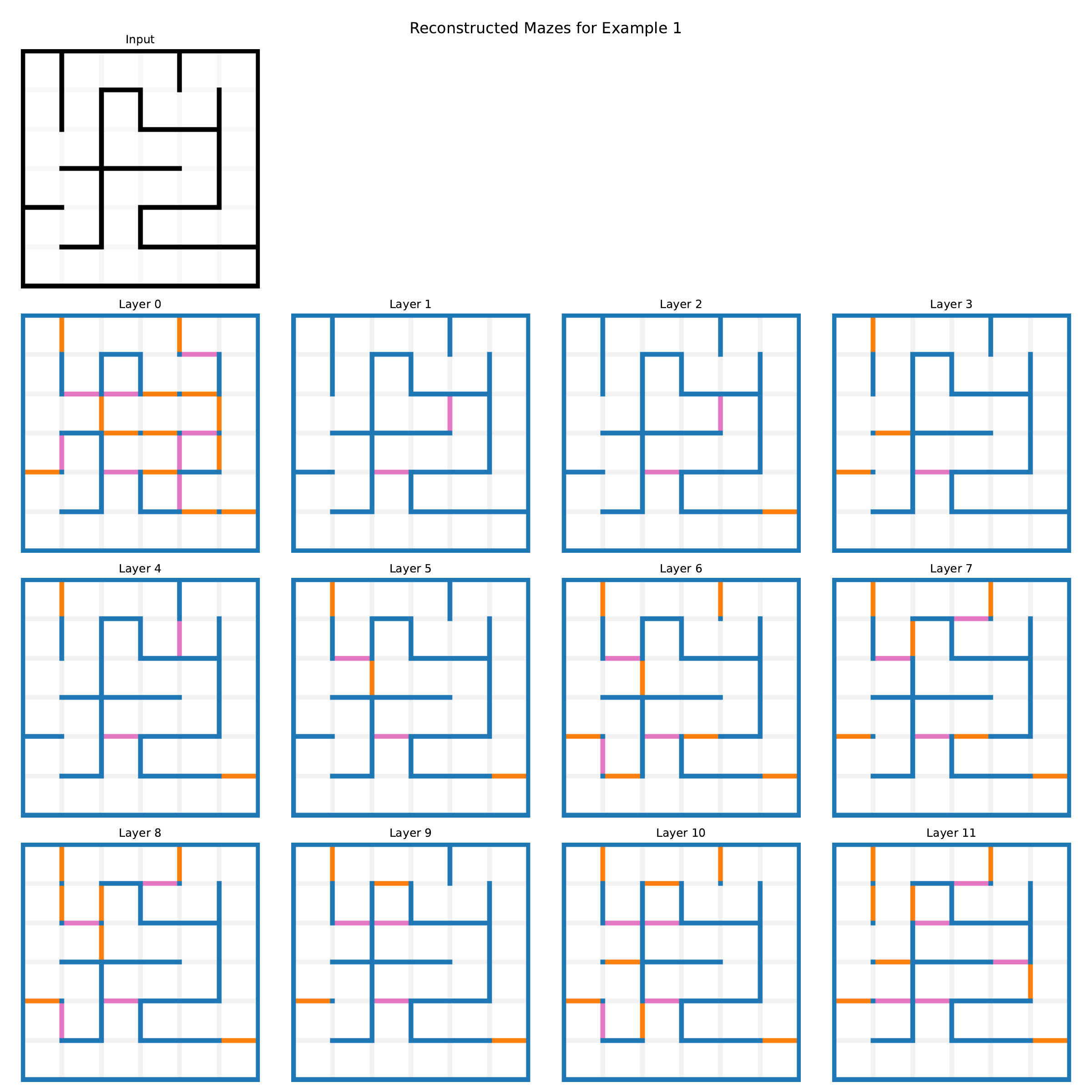}
    \caption{Decoding a maze with probe sets trained for each layer of the \jirpy\ model. Wall colors indicate that thresholded probes \textcolor{mpldarkblue}{Correctly Predicted}, \textcolor{mplorange}{Omitted} or \textcolor{mplpink}{Added} a wall.}
    \label{fig:app:reco_mazes_more_2}
\end{figure}

\end{document}